\documentclass[12pt]{article}
\usepackage{amsmath}
\usepackage{times}
\usepackage{graphicx}
\usepackage{color}
\usepackage{multirow}
\usepackage[authoryear]{natbib}
\usepackage{rotating}
\usepackage{bbm}
\usepackage{latexsym}

\usepackage{nicefrac}       %
\usepackage{tikz} %
\usepackage{floatrow}
\usepackage{placeins}
\usepackage{subcaption}
\usepackage{caption}

\usepackage{url}            %
\usepackage{booktabs}       %
\usepackage{amsfonts}       %
\usepackage{amsthm}
\usepackage{thmtools}
\usepackage{thm-restate}
\usepackage{microtype}      %
\usepackage{xcolor}         %

\newcommand{\captionfonts}{\normalsize}

\makeatletter  
\long\def\@makecaption#1#2{%
  \vskip\abovecaptionskip
  \sbox\@tempboxa{{\captionfonts #1: #2}}%
  \ifdim \wd\@tempboxa >\hsize
    {\captionfonts #1: #2\par}
  \else
    \hbox to\hsize{\hfil\box\@tempboxa\hfil}%
  \fi
  \vskip\belowcaptionskip}
\makeatother   

\newcommand{\bs}{\boldsymbol}

\declaretheorem[name=Lemma]{lemma}
\declaretheorem[name=Definition,numberwithin=section]{definition}

\usepackage{mathtools}

\newcommand{\pht}{\mathbf{x}_{t,a_t}^*}
\newcommand{\phtOpt}{\mathbf{x}_{t,a^*_t}^*}
\newcommand{\phtBar}{\mathbf{x}_{t, \bar a_t}^*}
\newcommand{\phs}{\mathbf{x}_{i,a_i}^*}
\newcommand{\phk}{\mathbf{x}_{t,k}^*}
\newcommand{\phra}{\mathbf{x}_{t,A_t}^*}

\newcommand{\pst}{\mathbf{x}_{t,a_t}}
\newcommand{\pstOpt}{\mathbf{x}_{t,a^*_t}}
\newcommand{\pstBar}{\mathbf{x}_{t, \bar a_t}}
\newcommand{\pss}{\mathbf{x}_{i,a_i}}
\newcommand{\psk}{\mathbf{x}_{t,k}}

\newcommand{\N}[2]{\mathcal{N}(#1,#2)}  
\newcommand{\eq}[1]{Eq.~(\ref{eq:#1})}

\newcommand{\cs}{Cauchy-Schwarz }
  
\newcommand{\fil}{\mathcal{F}}  
\newcommand{\rb}{\mathbb{R}}  
\newcommand{\eb}{\mathbb{E}}  
\newcommand{\ac}{\mathcal{A}} 
\newcommand{\ewt}{E^{\mathbf{w}}_t}
\newcommand{\ett}{E^\Theta_t}
\newcommand{\whct}{\mathbf{\widehat{w}}_t}
\newcommand{\wtct}{\Tilde{\mathbf{w}}_t}

\newcommand{\alignedintertext}[1]{%
  \noalign{%
    \vskip\belowdisplayshortskip
    \vtop{\hsize=\linewidth#1\par
    \expandafter}%
    \expandafter\prevdepth\the\prevdepth
  }%
}

\DeclareMathOperator*{\argmax}{argmax}

\begin{document}

\hspace{13.9cm}

\ \vspace{20mm}\\

{\LARGE Recurrent Neural-Linear Posterior Sampling for Non-Stationary Contextual Bandits}

\ \\
{\bf \large Aditya Ramesh}$^{\displaystyle * \ \displaystyle 1, \displaystyle 2, \displaystyle 3 }$\\
{\bf \large Paulo Rauber}$^{\displaystyle * \ \displaystyle 4 }$ \\
{\bf \large Michelangelo Conserva}$^{\displaystyle 4}$\\
{\bf \large Jürgen Schmidhuber}$^{\displaystyle 1, \displaystyle 2, \displaystyle 3, \displaystyle 5, \displaystyle 6}$\\
{$^{\displaystyle 1}$IDSIA. Lugano, Switzerland.}\\
{$^{\displaystyle 2}$USI. Lugano, Switzerland.}\\
{$^{\displaystyle 3}$SUPSI. Lugano, Switzerland.}\\
{$^{\displaystyle 4}$QMUL. London, United Kingdom.}\\
{$^{\displaystyle 5}$NNAISENSE. Lugano, Switzerland.}\\
{$^{\displaystyle 6}$KAUST. Thuwal, Saudi Arabia.}\\
{$^{\displaystyle *}$Equal contribution}\\


\thispagestyle{empty}
\markboth{}{NC instructions}
\ \vspace{-0mm}\\
\begin{center} {\bf Abstract} \end{center}
An agent in a non-stationary contextual bandit problem should balance between exploration and the exploitation of (periodic or structured) patterns present in its previous experiences. Handcrafting an appropriate historical context is an attractive alternative to transform a non-stationary problem into a stationary problem that can be solved efficiently. However, even a carefully designed historical context may introduce spurious relationships or lack a convenient representation of crucial information. In order to address these issues, we propose an approach that learns to represent the relevant context for a decision based solely on the raw history of interactions between the agent and the environment. This approach relies on a combination of features extracted by recurrent neural networks with a contextual linear bandit algorithm based on posterior sampling. Our experiments on a diverse selection of contextual and non-contextual non-stationary problems show that our recurrent approach consistently outperforms its feedforward counterpart, which requires handcrafted historical contexts, while being more widely applicable than conventional non-stationary bandit algorithms. Although it is very difficult to provide theoretical performance guarantees for our new approach, we also prove a novel regret bound for linear posterior sampling with measurement error that may serve as a foundation for future theoretical work.

\section{Introduction}
\label{sec:intro}

In a broad formulation of a contextual bandit problem, an agent chooses an arm (action) based on a context (observation) and previous interactions with an environment. In response, the environment transitions into a new hidden state and provides a reward and a new context. The goal of the agent is to maximize cumulative reward through a finite number of interactions with the environment, which requires balancing exploration and exploitation.

Many practical problems can be seen as contextual bandit problems \citep{bouneffouf2019survey}. For example, consider the problem of product recommendation: a context may encode information about an individual, an arm may represent a recommendation, and a reward may signal whether a recommendation succeeded.

If the expected reward is an (unknown) fixed linear function of a (known) vector that represents the preceding arm and context, independently of the remaining history of interactions between the agent and the environment, then several contextual linear bandit algorithms provide strong performance guarantees relative to the best fixed policy that maps contexts to arms \citep{auer2002using, abbasi2011improved, agrawal2013thompson, abeille2017linear}.

However, in a non-stationary contextual bandit problem, the state of the environment changes in such a way that the performance of any fixed policy that maps contexts to arms is unsatisfactory \citep{lattimore19bandit}. In the product recommendation example, the success rate of a recommendation may depend both on the time of the year and the results of previous recommendations. Therefore, the presence of this information in the contexts determines whether the problem is non-stationary.

Handcrafting an appropriate context is an attractive alternative to transform a non-stationary problem into a stationary problem that can be solved efficiently \citep{lattimore19bandit}. Unfortunately, an inappropriate context may introduce spurious relationships or lack a convenient representation of crucial information.

Another alternative is to employ non-stationary bandit algorithms, which can be divided into two main families  \citep{liu2018change, cao2019nearly,russac2020algorithms}. \emph{Passive} algorithms bias their decisions based on recent interactions with the environment, while \emph{active} algorithms attempt to detect when a significant change occurs. Unfortunately, algorithms from both families are incapable of exploiting periodicity and structure (the effect of actions on the rewards of future actions), which may be important even when no planning is required.

In order to address these issues, we propose an approach based on a recurrent neural network that receives the raw history of interactions between the agent and the environment. This network is trained to predict the reward for each pair of arm and context. The features extracted by the network are combined with a contextual linear bandit algorithm based on posterior sampling \citep{agrawal2013thompson}, which potentially allows an agent to achieve high performance in a non-stationary contextual problem without carefully handcrafted historical contexts. Besides its advantages in contextual problems, our approach is also radically different from previous approaches that are able to exploit periodic or structured patterns in non-contextual non-stationary bandit problems.

Our approach is partially motivated by the work of \citet{riquelme2018deep}, whose comprehensive experiments have shown that the combination of features extracted by a (feedforward) neural network with a contextual linear bandit algorithm based on posterior sampling achieves remarkable success in (stationary) contextual bandit problems. Our approach can also be seen as a model-based counterpart to recent model-free meta-learning approaches based on recurrent neural networks that have been applied to non-contextual stationary bandit problems \citep{duan2016rl, wang2016learning}.

We evaluate our approach using a diverse selection of contextual and non-contextual non-stationary bandit problems. The results of this evaluation show that our recurrent approach consistently outperforms its feedforward counterpart, which requires handcrafted historical contexts, while being more widely applicable than conventional non-stationary bandit algorithms. 
Despite this empirical success, our novel theoretical analysis of linear posterior sampling with measurement error suggests that our approach requires neural networks to generalize efficiently to guarantee good performance.

The remainder of this paper is organized as follows. Section \ref{sec:methodology} introduces our new approach after providing the necessary background. Section \ref{sec:theoretical_analysis} presents a novel theoretical analysis of linear posterior sampling with measurement error that provides some theoretical insight about the performance our new approach. Section \ref{sec:experiments} presents the results of an empirical comparison between our approach and alternative approaches. Finally, Section \ref{sec:conclusion} summarizes our findings and suggests future work.

\section{Methodology}
\label{sec:methodology}
\subsection{Preliminaries}

We denote random variables by upper case letters and assignments to these variables by corresponding lower case letters. We omit the subscript that typically relates a probability function to random variables when there is no risk of ambiguity. For example, we may use $p(x)$ to denote $p_{X}(x)$ in the same context where we use $p(y)$ to denote $p_{Y}(y)$.

\begin{figure}[ht]
\centering
    \begin{floatrow}[1]
      \ffigbox{\caption{Directed graphical models that represent the interaction between the agent or oracle and the environment. Dashed or dotted edges belong respectively to either the agent or the oracle.}\label{fig:pgm}}{%
        \includegraphics[width=0.6\textwidth]{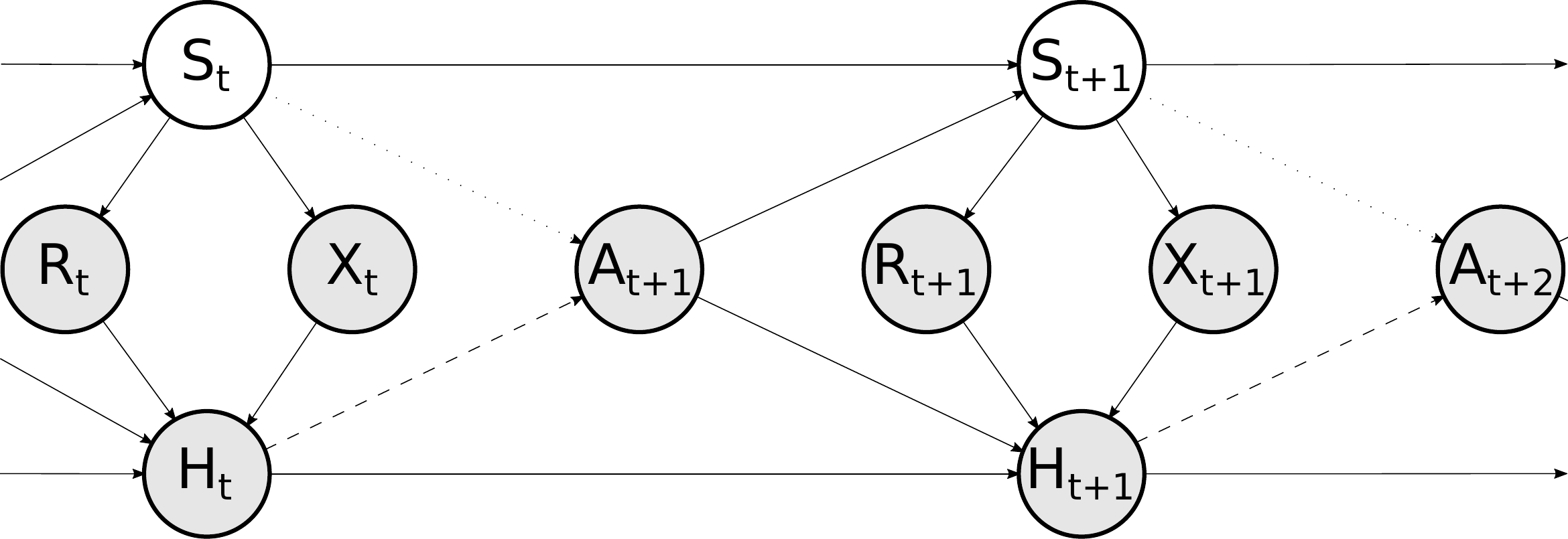}
      }
      \end{floatrow}
\end{figure}

A contextual bandit problem can be seen as a special case of the following partially observable reinforcement learning problem. An agent interacts with an environment (multi-armed bandit) during a single episode that lasts $T$ time steps. At a given time step $t \in [T] = \{1, 2, \ldots, T-1, T\}$, the environment is in a hidden state $S_t$, and the agent uses a policy $\pi$ to choose an action (arm) $A_{t+1}$ given the history $H_{t}$, which encodes the previous rewards $R_{1:t}$, observations (contexts) $X_{1:t}$, and actions $A_{2:t}$. In response to this action, the environment transitions into a hidden state $S_{t+1}$, and outputs a reward $R_{t+1}$ and an observation $X_{t+1}$. This process can be represented by the directed graphical model in Figure \ref{fig:pgm}.

In contrast to an agent, an \emph{oracle} uses a policy $\pi^*$ to choose an action $a_{t+1}$ that maximizes the immediate expected reward $\mathbb{E} \left[ R_{t+1} \mid s_{t}, a_{t+1} \right]$ given the hidden state $s_{t}$, for all $t$. Note that such oracle makes greedy decisions. Although non-greedy agents may achieve higher expected cumulative reward in fully fledged reinforcement learning environments, such environments are out of our scope.

The regret of a policy $\pi$ is given by $ \sum_{t=1}^T \mathbb{E}_{\pi^{*}}\left[  R_{t} \right] - \mathbb{E}_{\pi} \left[  R_{t} \right]$, where $\pi^*$ is an oracle policy, and the subscript on an expectation denotes the policy used for choosing actions. We are generally interested in policies that have low regret across a family of environments.

\subsection{Linear posterior sampling for contextual bandits}
\label{sec:posterior_sampling}

This section presents a decision-making algorithm for contextual linear bandits that is at the core of our proposed approach. \citet{agrawal2013thompson} were the first to provide strong theoretical guarantees for this algorithm under standard technical assumptions. 

Suppose that the expected reward for time step $t$ given the history $h_{t - 1}$, the action $a_{t}$, and an  (unknown) weight vector $\mathbf{w}$ is given by $\mathbb{E}\left[ R_{t} \mid h_{t - 1}, a_{t} , \mathbf{w} \right] = \mathbf{w} \cdot \bs{\phi}(x_{t-1}, a_{t})$, where the feature map $\bs{\phi}$ is a (known) function responsible for encoding any given pair of observation and action into a feature vector. In other words, suppose that the expected reward for a given time step is an unknown linear function of a known feature vector that represents the previous observation and the chosen action, independently of the rest of the history.

In this setting, posterior sampling starts by representing knowledge about $\mathbf{W}$ in a prior distribution. At a given time step $t$, the algorithm consists of four simple steps: (1) drawing a single parameter vector $\mathbf{w}_t$ from the prior over $\mathbf{W}$; (2) choosing an action $a_{t}$ that maximizes $\mathbf{w}_{t} \cdot \bs{\phi}(x_{t-1}, a_{t})$, (3) observing the reward $r_{t}$; (4) computing the posterior over $\mathbf{W}$ to be used as a prior for step $t+1$. Intuitively, at a given time step, an action is drawn according to the probability that it is optimal.

In order to derive an efficient algorithm, suppose that the prior density for $\mathbf{w}$ is given by $p(\mathbf{w}) = \mathcal{N}(\mathbf{w} \mid \mathbf{w}_0, \mathbf{V}_0)$, for some hyperparameters $\mathbf{w}_0$ and $\mathbf{V}_0$.

Furthermore, consider the dataset $\mathcal{D} = \{ (\bs{\phi}(x_{t' - 1}, a_{t'}), r_{t'}) \}_{t' = 2}^t$, and suppose that the conditional likelihood of the parameter vector $\mathbf{w}$ is given by $p(\mathcal{D} \mid \mathbf{w}) = \mathcal{N}(\mathbf{r} \mid \bs{\Phi}\mathbf{w}, \sigma^2 \mathbf{I})$, where $\mathbf{r} = (r_{2}, \ldots, r_{t})$ is the reward vector, $\bs{\Phi}$ is the design matrix where each row corresponds to a feature vector in $\mathcal{D}$, $\mathbf{I}$ is the appropriate identity matrix, and $\sigma^2 > 0$ is a hyperparameter.

In that case, the posterior density for $\mathbf{w}$ is given by $p(\mathbf{w} \mid \mathcal{D}) \propto_{\mathbf{w}} \mathcal{N}(\mathbf{r} \mid \mathbf{\bs{\Phi}}\mathbf{w}, \sigma^2\mathbf{I}) \mathcal{N}(\mathbf{w} \mid \mathbf{w}_0, \mathbf{V}_0)$. Because the random vectors $\mathbf{W}$ and $\mathbf{R}$ are related by a linear Gaussian system \citep{bishop2013pattern}, the desired posterior density is given by $p(\mathbf{w} \mid \mathcal{D}) = \mathcal{N}(\mathbf{w} \mid \bs{\mu}, \bs{\Sigma})$,
where
\begin{align}
 \bs{\Sigma}^{-1} =  \mathbf{V}_0^{-1} + \frac{1}{\sigma^2}\mathbf{\bs{\Phi}}^T\mathbf{\bs{\Phi}},\\
 \bs{\mu} = \frac{1}{\sigma^2}\bs{\Sigma}\mathbf{\bs{\Phi}}^T \mathbf{r} + \bs{\Sigma}\mathbf{V}_0^{-1}\mathbf{w}_0.
\end{align}

At a given time step, it is straightforward to draw a parameter vector from this multivariate Gaussian posterior density function (Step 1), choose the best corresponding action (Step 2), observe the outcome (Step 3), and update the dataset and the posterior (Step 4), which completes the algorithm.

Crucially, the assumptions of a multivariate Gaussian prior and a multivariate Gaussian likelihood are only used to derive an efficient algorithm. The conditions under which this algorithm achieves its theoretical guarantees are very permissive and somewhat unrelated \citep{agrawal2013thompson, abeille2017linear}. This is important because the dataset $\mathcal{D}$ is generally not composed of independent and identically distributed sample elements.

\subsection{Feedforward neural-linear feature vectors}

This section presents the process of extracting feedforward neural-linear feature vectors that ideally allow predicting the expected reward from any pair of history and action. A comprehensive benchmark has shown that the combination of these feature vectors with posterior sampling for contextual linear bandits often outperforms other posterior sampling approaches for contextual bandits \citep{riquelme2018deep}.

Consider the dataset $\mathcal{D} = \{ (\bs{\psi}(h_{t' - 1}, a_{t'}), r_{t'}) \}_{t'=2}^{t}$, where the feature map $\bs{\psi}$ is a function responsible for encoding the information that ideally allows predicting the reward $r_{t'}$ from the history $h_{t' - 1}$ and the action $a_{t'}$ into a feature vector, for all $t'$. In contrast to the previous section, we do not assume that the expected reward is a linear function of the corresponding feature vector.

In a stationary contextual problem, $\bs{\psi}(h_{t' - 1}, a_{t'})$ may encode just the observation $x_{t' - 1}$ and the action $a_{t'}$ in order to enable predicting $r_{t'}$. In a non-stationary contextual problem, $\bs{\psi}(h_{t' - 1}, a_{t'})$ may encode a (periodic function of) the current time step $t'$; statistics regarding actions; statistics regarding observations; the last $n$ rewards, observations, and actions; and arbitrary combinations of similar information. As will become clear, the need to handcraft an appropriate feature map $\bs{\psi}$ for a specific non-stationary problem is a potential weakness, since $\bs{\psi}$ may introduce spurious relationships or dismiss crucial information.

Extracting feedforward neural-linear feature vectors requires fitting a feedforward neural network to the dataset $\mathcal{D}$, which may be accomplished by searching for parameters that minimize a cost function using typical methods. Note that such methods assume that the dataset $\mathcal{D}$ is composed of independent and identically distributed sample elements, which is generally not the case, as in the previous section.

\begin{figure}[t]
    \begin{floatrow}[2]
      \ffigbox{\caption{Feedforward neural-linear network.}\label{fig:nn}}{%
      \includegraphics[height=0.1\textheight]{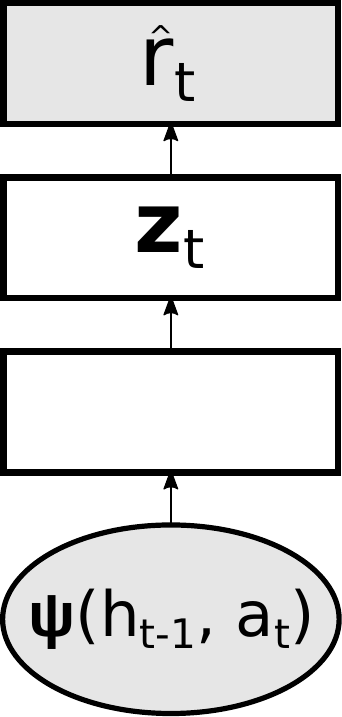}
      }
      \ffigbox{\caption{Recurrent neural-linear network.}\label{fig:rnn}}{%
        \includegraphics[height=0.1\textheight]{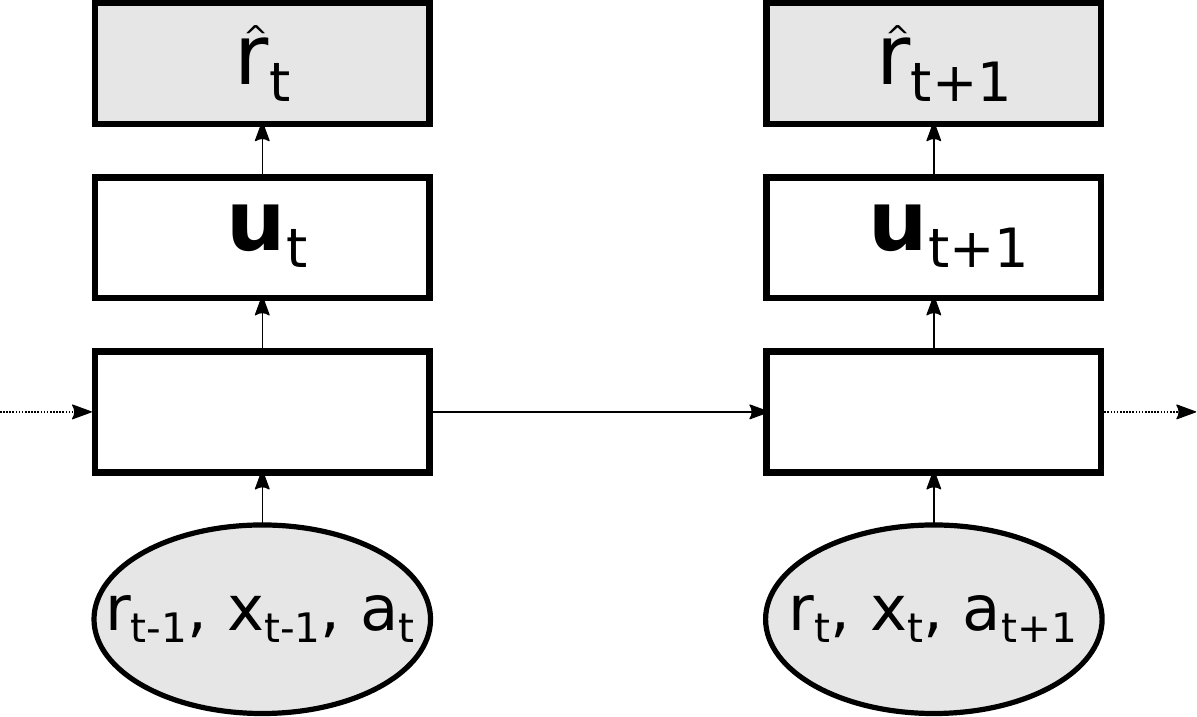}
      }
      \end{floatrow}
\end{figure}

The feedforward neural-linear feature vector $\mathbf{z}_{t'}$ is the output of the penultimate layer (last hidden layer) of the (fitted) neural network when given $\bs{\psi}(h_{t' - 1}, a_{t'})$ as input (Fig. \ref{fig:nn}). The only restriction on the network architecture is that the last layer should have a single linear unit (with no bias).
By construction, if the parameters of the neural network achieve low cost on the training dataset $\mathcal{D}$, then it should be possible to approximate the reward $r_{t'}$ as a linear function of the feedforward neural-linear feature vector $\mathbf{z}_{t'}$, for any $t' \leq t$. Under the strong assumption that this is also true for $t' > t$, feedforward neural-linear feature vectors can be combined with posterior sampling for contextual linear bandits to provide a complete algorithm for contextual bandits. Despite its lack of known theoretical guarantees, this algorithm excels experimentally \citep{riquelme2018deep}.

\subsection{Recurrent neural-linear feature vectors}

This section introduces the novel process of extracting recurrent neural-linear feature vectors that ideally allow predicting the expected reward from any pair of history and action. In contrast to the feedforward approach, this process eliminates the need for a handcrafted feature map $\bs{\psi}$.

Consider the dataset $\mathcal{D} = \{ ( \tau_{t'}, r_{t'})  \}_{t' = 2}^t$, where $\tau_{t'} = (r_{1}, x_{1}, a_{2}, \ldots, r_{t'- 1}, x_{t'-1}, a_{t'})$ represents the interaction between the agent and the environment up to time step $t'$.

Extracting recurrent neural-linear feature vectors requires fitting a recurrent neural network to the dataset $\mathcal{D}$, which may be accomplished by searching for parameters that minimize a cost function using typical methods. At a given time step $t'$, this network receives as input the reward $r_{t'-1}$, the observation $x_{t'-1}$, the action $a_{t'}$, and attempts to predict the reward $r_{t'}$ (Fig. \ref{fig:rnn}).

The recurrent neural-linear feature vector $\mathbf{u}_{t'}$ is the output of the penultimate layer of the (fitted) recurrent neural network when given $\tau_{t'}$ as input. As in the previous section, the only restriction placed on the network architecture is that the last layer should have a single linear unit (with no bias).

As in the feedforward approach, if the parameters of the recurrent neural network achieve low cost on the training dataset $\mathcal{D}$, then it should be possible to approximate the reward $r_{t'}$ as a linear function of the recurrent neural-linear feature vector $\mathbf{u}_{t'}$, for any $t' \leq t$. Under the strong assumption that this is also true for $t' > t$, recurrent neural-linear feature vectors can be combined with posterior sampling for contextual linear bandits to provide a complete algorithm for contextual bandits. 

Most importantly, the recurrent neural-linear approach eliminates the need for a handcrafted feature map $\bs{\psi}$. Besides its potential advantages in contextual problems, this approach is also radically different from previous approaches that are able to exploit periodic or structured patterns in non-contextual problems.

\section{Theoretical analysis}
\label{sec:theoretical_analysis}

The suitability of (recurrent or feedforward) neural-linear feature vectors as contexts for linear posterior sampling is entirely dependent upon the capacity of neural networks to generalize.
In this section, we present our novel theoretical analysis of linear posterior sampling with measurement error. This analysis can be safely skipped by readers who are mostly interested in empirical results.

Our regret analysis is based on previous theoretical work on linear posterior sampling \citep{agrawal2012thompson} and its neural counterpart \citep{zhang2021neural}.
In contrast with the typical linear posterior sampling setting \citep{agrawal2012thompson}, the measurement error considered in our setting leads to a biased estimation procedure. The measurement error can be related to the neural network approximation error considered by \citet{zhang2021neural}.
However, their analysis addresses a deep neural network variant of posterior sampling that is significantly different from linear posterior sampling.

Section \ref{sec:theoretical_analysis:preliminaries} introduces the linear contextual bandit with measurement error setting and a slightly modified version of our notation to facilitate the exposition.
Section \ref{sec:theoretical_analysis:theorem} presents the main outcome of our regret analysis (Thm.~\ref{th:regret}) along with its proof.
Appendices  \ref{app:theoretical_analysis:lemma1}, \ref{app:theoretical_analysis:lemma2}, \ref{app:theoretical_analysis:lemma3}, and
\ref{app:theoretical_analysis:lemma4}
contain the Lemmas required to prove Theorem \ref{th:regret}.

\subsection{Preliminaries} \label{sec:theoretical_analysis:preliminaries}

We let $[N]$ denote the set $\{1,\ldots,N\}$ for any positive integer $N$.
If $\mathbf{x}$ is a real vector, its Euclidean norm is denoted by $\|\mathbf{x}\|$, and its matrix weighted norm is denoted by $\|\mathbf{x}\|_ A = \sqrt{\mathbf{x}^T A \mathbf{x}}\,$, where $A$ is a positive semidefinite matrix. The inner product is denoted by $\cdot$ and the outer product by $\otimes$.
The indicator function is denoted by $\mathcal{I}$.

An agent interacts with an environment during $T$ time steps. 
At every time step $t \in [T]$ and for each action $k \in[K]$, the agent is presented with a context with measurement error $\psk\in\rb^d$ that corresponds to a true but \emph{unobserved} context $\phk\in\rb^d$. Based on the \emph{observed} context $\psk$, the agent chooses an action $A_t \in[K]$.
In response to this action, the environment generates a reward $R_{t,A_t} = \mathbf{w} \cdot \phra + \eta_t$, where $\mathbf{w} \in \rb^d$ is the environment parameter and $\eta_t$ is additive noise. Note that the environment generates rewards using the \emph{true} contexts.

We formalize the interaction between the agent and the environment as a stochastic process corresponding to the filtration
$\fil_t$ 
= 
$\{
\mathcal{H}_t, 
\{\mathbf{x}_{t,k}\}_{k=1}^K
\}
$,
where
\begin{equation}
    \mathcal{H}_t = 
(
\{ \mathbf{x}_{1,k}\}_{k=1}^K,
A_1, 
R_{1,A_1}
,
\ldots
,
\{ \mathbf{x}_{t-1,k}\}_{k=1}^K, 
A_{t-1}, 
R_{t-1,A_{t-1}}
)
\end{equation}
is the history encoding the previous actions, contexts, and rewards.
We assume that the true context vectors are chosen adaptively by an adversary and that the measurement error is, instead, independent.%

The goal of the agent is to minimize the regret during the interaction with the environment.
Formally, the cumulative regret after $T$ time steps is $\sum_{t=1}^T \texttt{regret}_{t,A_t}$, where ${\texttt{regret}_{t,k} = \mathbf{w} \cdot \phtOpt - \mathbf{w} \cdot \phk}$ is the instantaneous regret suffered by the agent at time step $t$ for action $k$ and ${a_t^* = \argmax_{k'} \mathbf{w} \cdot \mathbf{x}^*_{t,k'}}$ is the optimal arm.
Note that we can apply a similar procedure to the one described in Appendix A.5 of \citet{agrawal2012thompson} to extend our results for the case in which the regret is alternatively defined as $\texttt{regret}_t = R_{t, a_t^*} - R_{t, A_t}$.

In our regret analysis, we assume that the reward noise $\eta_t$ is conditionally $\sigma$-sub-Gaussian for $\sigma\geq 0$ (see Def.~\ref{definition:subG}).
This assumption is weak since it is always satisfied when the reward is bounded. 
In order to obtain a scale-free regret bound, we also assume that, for all $t \in [T]$ and $k \in [K]$, $\|\mathbf{w}\|\leq1$, $\|\psk\|\leq1$, and $\|\phk\|\leq1$.
As typical in the measurement error literature, we assume that $\psk$ lies in the same space as $\phk$.
This assumption may not hold in the case of neural-linear features.
However, our proof can be easily extended to a setting with different dimensionalities.
Finally, we assume that the measurement error is bounded, satisfying $\|\psk - \phk \| \leq \epsilon_t$ for all $t \in [T]$  and $k \in[K]$.
Since we do not want to make assumptions on the functional form of $\epsilon_t$, our analysis is presented in terms of $\epsilon$ such that $\epsilon_t \leq \epsilon$ for all $t \in [T]$.

In the following, we introduce the version of linear posterior sampling considered in the proof. This version closely resembles the one presented by \citet{agrawal2012thompson}, except for the measurement error. Section \ref{sec:posterior_sampling} contains a higher-level description of the same algorithm.

We assume a Gaussian prior and a Gaussian likelihood for the reward to efficiently compute the posterior. Note that this assumption is only required to derive an efficient algorithm. The only assumption regarding the reward generation process is of $\sigma$-sub-Gaussianity.

Formally, we suppose that the conditional likelihood of the reward for each action $k \in [K]$ is normally distributed as

\begin{equation}
    R_{t,k}|\fil_t \sim \N{\mathbf{w} \cdot \phk}{\nu_t^2}, 
    \,\,\,
    \text{where} 
    \,\,
    \nu_t=3 \sigma \sqrt{d \log \left ( \frac{t}{\delta} \right )} + \epsilon \sqrt{t d \log \left ( \frac{t+d}{d} \right ) }.
\end{equation}
Note that $\nu_t$ depends on $\epsilon$. 
This corresponds to inflating the variance, which is equivalent to accounting for the measurement error.
From a practical perspective, $\epsilon$ is typically unknown but can be treated as a hyperparameter.

If we let $\N{\whct}{\nu_t^2\mathbf{B}_t^{-1}}$ be the prior at timestep $t$ for the environment parameter $\mathbf{w}$ given $\fil_t$, then 
$\N{\mathbf{\widehat{w}}_{t+1}}{\nu_{t+1}^2\mathbf{B}_{t+1}^{-1}}$ is the posterior at the next timestep $t+1$ given $ R_{t,A_t}, \fil_{t+1}$, where
\begin{equation}
    \mathbf{B}_t = \mathbf{I}_d + \sum_{i=1}^{t-1} \pss \otimes \pss
    \hspace{.4cm}
    \text{and}
    \hspace{.4cm}
    \whct = \mathbf{B}_t^{-1} \sum_{i=1}^{t-1} \pss r_{i,a_i}.
\end{equation}

At each time step $t$, the agent generates a sample ${\wtct}$ from the posterior $\N{\whct}{\nu_t^2\mathbf{B}_t^{-1}}$ over the environment parameter
and chooses the optimal action for $\wtct$. 
Concretely, the algorithm chooses ${A_t = \argmax_{k} \Theta_{t,k}}$, breaking ties randomly, where $\Theta_{t,k} = \wtct \cdot \psk$.
It follows that $\Theta_{t,k}$ is distributed as $\N{\whct \cdot \psk}{\nu_t^2s_{t,k}^2}$,
where $s_{t,k} = \|\psk\|_{\mathbf{B}_t^{-1}}$.

Following \citet{agrawal2012thompson} and \citet{zhang2021neural}, we divide the arms into two groups: saturated and unsaturated. The optimal arm is in the group of unsaturated arms. Intuitively, the unsaturated arms are those worth selecting.
Formally, the set of saturated arms $C_t$ is given by
\begin{equation}
    C_t = \left \{ k \in [K] :  \phtOpt \cdot \mathbf{w} - \phk \cdot \mathbf{w} > s_{t,k} g_t + 2\epsilon \right \}, 
\end{equation}
where $g_t = \min \{2\sqrt{d\log(t)},2\sqrt{\log(tK)}\}\nu_t + l_t$,
and $l_t = \sigma \sqrt{d\log\left(\frac{t^3}{\delta}\right)}+ 1 + \\ \epsilon \sqrt{t d \log \left ( \frac{t+d}{d} \right ) }$.
\begin{align}
\intertext{The event $\ewt$ is that $\whct \cdot \psk$ suitably concentrates around the expected value of the reward is given by}
    E^{\mathbf{w}}_t =\,& \left \{\omega \in \fil_{t} : \forall\,k \in [K], \quad 
    |\psk \cdot \whct - \phk \cdot \mathbf{w}| \leq l_t s_{t,k} +\epsilon \right \}.
\intertext{The event $\ett$ is that $\Theta_{t,k}$ suitably concentrates around its mean is given by}
    E^{\Theta}_t =\,& \left \{\omega \in \fil_{t} : \forall\,k \in [K], 
       |\Theta_{t,k} - \psk \cdot \whct| \leq \nu_t s_{t,k} \min\{2\sqrt{d\log(t)},2\sqrt{\log(tK)}\} \right \}.
\end{align}

\subsection{Regret of linear posterior sampling with measurement error}
\label{sec:theoretical_analysis:theorem}

The following theorem is the main outcome of our analysis.

\begin{restatable}[Regret of linear posterior sampling with measurement error]{theorem}{theoremregret}
Let $\psk$ be a context with measurement error corresponding to the true context $\phk$ at time step $t$ for action $k$.
Suppose that the measurement error satisfies $\|\psk - \phk \| \leq \epsilon_t$ for all $t \in [T]$, $k \in [K]$, and some $\epsilon_t \in \rb^+$.
For any $\epsilon \geq \epsilon_t$, under the standard assumptions described in Section~\ref{sec:theoretical_analysis:preliminaries}, the cumulative regret of linear posterior sampling is bounded as
$$
\widetilde{O} \left(
d^{\frac{3}{2}} \sqrt{T} + \epsilon d^{\frac{3}{2}} T
\right)
\quad
\text{or}
\quad
\widetilde{O}
\left(
d\sqrt{T \log(K)} + \epsilon d \sqrt{\log(K)} T
\right),$$
whichever is smaller, with probability $1-\delta$.
\label{th:regret}
\end{restatable}

\textit{Proof outline.}\hspace{1mm}
Under the assumption that $\ewt$ holds,
we construct a stochastic process $\big(Y_t; t\geq0 \big)$ related to the cumulative regret.
By Lemma~\ref{th:lemma4}, $Y_t$ is a super-martingale, and so we can use the Azuma-Hoeffding inequality (see Def. \ref{lemma:azuma_hoeffding}) to bound it with high probability.
Finally, we eliminate the dependency on $\ewt$ by taking a union bound over the high probability bound from the Azuma-Hoeffding inequality and Lemma~\ref{th:lemma1}, which bounds the probability of the event $\ewt$.

\begin{proof}
Let $\big(Y_t = \sum_{i=1}^t Z_i; t\geq0 \big)$ be a stochastic process corresponding to $\fil_t$ where
\begin{equation}
    \texttt{regret}'_{t,A_t} = \texttt{regret}_{t,A_t} \mathcal{I} \left \{\ewt \right \}
    \hspace{.2cm}
    \text{and}
    \hspace{.2cm}
    Z_t = \texttt{regret}'_{t,A_t} - 4\epsilon - 44 e \sqrt{\pi}g_t s_{t,A_t} - \frac{2}{t^2}.
\end{equation}
Recall that the instantaneous regret at $t$ is defined as ${\texttt{regret}_{t,A_t} = \mathbf{w} \cdot \phtOpt - \mathbf{w} \cdot \phra}$, where ${a_t^* = \argmax_{k'} \mathbf{w} \cdot \mathbf{x}^*_{t,k'}}$ is the optimal arm.

By Lemma~\ref{th:lemma4}, $Y_t$ forms a super-martingale process as
\begin{equation*}
   \eb \left( Y_t - Y_{t-1} \right) = \eb \left(Z_t\right) \leq 4 + 4\epsilon + 44e\sqrt{\pi}g_T = K_T.
\end{equation*}
Using the Azuma-Hoeffding inequality for super-martingale processes,
\begin{equation}\label{eq:t2}
\frac{Y_T - Y_0}{K_T} \leq \sqrt{2T\log \left ( \frac{2}{\delta} \right )}  
\end{equation}
with $1-\frac{\delta}{2}$ probability.

Taking the union bound over Lemma~\ref{th:lemma1} and \eq{t2}, we can bound the regret as
\begin{align} \label{eq:regret_final}
\sum_{t=1}^T \texttt{regret}_{t,A_t} \leq & 4\epsilon T + 44 e \sqrt{\pi}g_T 5\sqrt{dT\log T} + \frac{\pi^2}{3} \notag \\ 
& + \left (4 + 4\epsilon + 44e\sqrt{\pi}g_T \right )\sqrt{2T\log \left ( \frac{2}{\delta} \right )}
\end{align}
 with $1-\delta$ probability. The asymptotic result can be obtained by observing that $g_T$ is in $\widetilde{O} \left(
d + \epsilon d \sqrt{T}
\right)$ or $\widetilde{O} \left(
\sqrt{d \log(K)} + \epsilon \sqrt{dT \log(K)}
\right)$, whichever term is smaller. The complete analysis is provided in Appendix \ref{app:theoretical_analysis:regret_theorem}.
\end{proof}

Unless the upper bound in Theorem \ref{th:regret} can be decreased significantly, this result suggests that the distance between the neural-linear feature vectors and idealized context vectors must decrease quickly in order for (recurrent or feedforward) neural-linear posterior sampling to perform generally well. This requires efficient generalization from neural networks.

\section{Experiments}
\label{sec:experiments}

This section reports results of an empirical comparison between our recurrent neural-linear approach, its feedforward counterpart, and conventional non-stationary bandit algorithms. We also study the impact of neural-linear posterior sampling on our recurrent approach by replacing it with simpler exploration strategies.
\subsection{Bandit problems}
\label{sec:experiments:bandit_problems}

We performed experiments on a diverse selection of contextual and non-contextual non-stationary bandit problems, which is detailed below. %
Because there are no standard benchmarks for non-stationary bandit algorithms, this selection combines original problems with problems borrowed from previous work. 
These problems may be partitioned into four categories according to their underlying non-stationarity: abrupt periodic, smooth periodic, structured (where an action may affect the rewards of future actions), or unknown (derived from a real dataset).

\subsubsection{Non-contextual bandit problems} 
\label{sec:experiments:bandit_problems:non_contextual}

In the (abrupt periodic) \emph{flipping Gaussian} and \emph{flipping Bernoulli} problems, the mean reward of each arm switches abruptly every fixed number of time steps. In the (smooth periodic) \emph{sinusoidal Bernoulli} problem, the mean reward of each arm is a sinusoidal function of the current time step. In the (structured) \emph{circular Markov chain} problem, the best arm trades place with the next arm in a pre-defined cyclical order after it is found.

The configurations of the four non-stationary non-contextual bandit problems are described below.

\paragraph{Flipping Gaussian.} The mean of the Gaussian reward for each arm $k$ changes from $\mu_k$ to $-\mu_k$ every $h$ time steps, while the corresponding variance $s^2$ is fixed across arms. We chose $K=8$ arms, $h=10$, initial means in $\{ 0.1, 0.2, \ldots, 0.9 \} \setminus \{ 0.5\}$, variance $s^2 = 0.1^2$.

\paragraph{Flipping Bernoulli.} The mean of the Bernoulli reward for each arm $k$ changes from $p_k$ to $1 - p_k$ every $h$ time steps. We chose $K=8$ arms, $h=10$, initial means in $\{ 0.1, 0.2, \ldots, 0.9 \} \setminus \{ 0.5\}$.

\paragraph{Sinusoidal Bernoulli.} The mean of the Bernoulli reward for each arm is a sinusoidal function of the current time step. The frequency of each function is the same across arms, but the phase is different. Concretely, the mean of the Bernoulli reward for each arm $k$ at time step $t$ is given by $p_k = 1/2 + \sin[ 2\pi f t + 2\pi(k-1)/K ]/2$, where $K=5$ is the number of arms and $f = 1/32$ is the frequency. This environment enables comparing our proposed approach with more conventional algorithms such as discounted UCB and sliding-window UCB, which were previously compared in a similar (albeit much simpler) environment \citep{garivier2011upper}.

\paragraph{Circular Markov chain.} The mean of the Gaussian reward for every arm is $\mu$, with corresponding variance $s^2$, except for a single arm whose Gaussian reward has a mean $\mu^* > \mu$. After this arm is chosen, it trades place with the next arm in a predefined cyclical order. Note that an action may affect the reward of other actions in the future. We chose $K=8$ arms, common mean $\mu = 0$, best mean $\mu^* = 1$, and variance $s^2 = 0.05^2$.

\subsubsection{Contextual bandit problems} 
\label{sec:experiments:bandit_problems:contextual}

In the (abrupt periodic) \emph{flipping digits} problem, each context corresponds to an image of a digit, and the best arm for each digit switches every fixed number of time steps. In the (unknown) \emph{wall-following robot} problem, each context encodes readings of sensors from a real robot, and the best arm depends on the underlying movement pattern of the robot. The remaining two problems are non-stationary contextual linear bandit problems. In the (abrupt periodic) \emph{flipping vector} problem, the expected reward measures the alignment between an action-dependent vector and a vector that switches direction every fixed number of time steps. In the (smooth periodic) \emph{rotating vector} problem, the expected reward measures the alignment between an action-dependent vector and a vector rotating about the origin. 

We performed experiments on the non-stationary contextual bandit problems described below. 

\paragraph{Flipping digits.} Each of the ten arms is labeled with a different digit. Each observation corresponds to an image of a digit from a subset of the MNIST dataset \citep{lecun2010mnist}. Initially, the mean of the Gaussian reward for every arm is $\mu$, with corresponding variance $s^2$, except for the arm that is labeled with the digit depicted in last observation, whose mean is $\mu^* > \mu$. Every $h$ time steps, the arm labeled with digit $k$ becomes labeled with digit $9-k$. We chose $h=64$, common mean $\mu=0$, best mean $\mu^*=1$, and variance $s^2 = 0.05^2$.

\paragraph{Wall-following robot.} This problem is derived from a sequential classification dataset \citep{dua2017, Friere2009wallfollowing}. The observation $x_{t-1}$ for time step $t-1$ encodes readings of $24$ sensors from a mobile robot. Each of the arms corresponds to one of four recognized movement patterns (forward, right turn, sharp right turn, left turn). The mean of the Gaussian reward for every arm is $\mu$, with corresponding variance $s^2$, except for the arm that corresponds to the current movement pattern. Identifying a movement pattern may require combining observations across time steps. We chose common mean $\mu=0$, best mean $\mu^* = 1$, and variance $s^2 = 0.05^2$.

The two problems described below are non-stationary contextual linear bandit problems. These problems require a slightly modified implementation, which is detailed in Section \ref{sec:implementation}.

\paragraph{Flipping vector.} The observation $x_{t-1}$ for time step $t-1$ encodes a vector $\mathbf{x}_{t-1,k}$ for each action $k$. Each of these vectors is drawn from a finite set of unit vectors in $\mathbb{R}^d$, without replacement within a time step. The probability density function for the reward at time step $t$ given the history $h_{t - 1}$, the action $a_{t}$, and a parameter vector $\mathbf{w}_t$ is given by $p(r_t \mid h_{t-1}, a_{t}, \mathbf{w}_t) = \mathcal{N}(r_t \mid \mathbf{w}_t \cdot \mathbf{x}_{t-1,a_{t}}, s^2)$, where $s^2 > 0$ is a variance. The parameter vector $\mathbf{w}_2$ is a randomly chosen unit vector. Every $h$ time steps, the parameter vector changes from $\mathbf{w}_{t}$ to $-\mathbf{w}_{t}$. In simple terms, the expected reward measures the alignment between an action-dependent vector and a vector that changes direction every $h$ time steps. We chose $K=25$ arms, $h = 64$, dimension $d=50$, and variance $s^2 = 0.05^2$.

\paragraph{Rotating vector.} The observation $x_{t-1}$ for time step $t-1$ encodes a vector $\mathbf{x}_{t-1,k}$ for each action $k$. Each of these vectors is drawn from a finite set of unit vectors in $\mathbb{R}^2$, without replacement within a time step. The probability density function for the reward at time step $t$ given the history $h_{t - 1}$, the action $a_{t}$, and a parameter vector $\mathbf{w}_t$ is given by $p(r_t \mid h_{t-1}, a_{t}, \mathbf{w}_t) = \mathcal{N}(r_t \mid \mathbf{w}_t \cdot \mathbf{x}_{t-1,a_{t}}, s^2)$, where $s^2 > 0$ is a variance. The parameter vector $\mathbf{w}_t$ is given by $\mathbf{w}_t = (\cos(2\pi f t), \sin(2\pi f t))$, where $f$ is a frequency. In simple terms, the expected reward measures the alignment between an action-dependent vector and a vector rotating about the origin. We chose $K=25$ arms, variance $s^2 = 0.05^2$, and two different frequencies ($f = 1/32$ or $f = 1/2048$) depending on the experiment. This problem is similar to a problem employed by \citet{russac2019dlinucb}, which enables comparing our proposed approach with more conventional algorithms such as discounted linear UCB \citep{russac2019dlinucb} and sliding-window linear UCB \citep{cheung2019swlinucb}.

\subsubsection{Experiments based on the Criteo dataset}
\label{sec:experiments:bandit_problems:criteo}

Using hyperparameters found to perform well in the environments listed in Section \ref{sec:experiments:bandit_problems:contextual}, we also present the results of experiments conducted with real-world advertising data from the Criteo dataset \citep{diemert2017attribution}.

We further consider two non-stationary contextual linear bandit problems, where the linear bandit parameter ($\mathbf{w}_t$) and action encodings ($\mathbf{x}_{t-1,k}$ for $k=1 \dots K$) are estimated from the Criteo dataset \citep{diemert2017attribution}.
Each row of the dataset corresponds to one impression (a banner) that was displayed to a user. For each banner, there are anonymized contextual features (\emph{cat1} to \emph{cat9}) and potential target variables such as whether the banner was clicked or not. For both non-stationary problems, we followed the same pre-processing steps used by \cite{kim2020randomized}. Similar steps are also used by \cite{russac2019dlinucb}. The variable \emph{campaign} and categorical variables \emph{cat1} to \emph{cat9} (except \emph{cat7}) are selected as contextual features. From the one-hot encoded contextual variables, $d$ features were selected using Singular Value Decomposition. The initial parameter vector $\mathbf{w}_2$ is obtained by fitting a linear regression model of these features to the target variable (click/no click). The observation $x_{t-1}$ for time step $t-1$ encodes a vector $\mathbf{x}_{t-1,k}$ for each action $k$. At each time step, the policy is presented with $\frac{K}{2}$ action encodings sampled from a pool of clicked banners and $\frac{K}{2}$ action encodings sampled from a pool of not clicked banners. The probability density function for the reward at time step $t$ given the history $h_{t - 1}$, the action $a_{t}$, and a parameter vector $\mathbf{w}_t$ is given by $p(r_t \mid h_{t-1}, a_{t}, \mathbf{w}_t) = \mathcal{N}(r_t \mid \mathbf{w}_t \cdot \mathbf{x}_{t-1,a_{t}}, s^2)$, where $s^2 > 0$ is a variance.  We chose $K=10$ arms, dimension $d=50$, and variance $s^2 = 0.15$.

The two considered sources of non-stationarity in the experiments with the Criteo dataset are described below.

\paragraph{Abrupt partial flip vector based on the Criteo dataset.}{A single abrupt partial flip occurs at $t=4000$. The signs of 60\% of the components of the initial linear parameter $\mathbf{w_2}$ are switched at the specific change point.}

\paragraph{Flipping vector based on the Criteo dataset.}{The parameter vector changes from $\mathbf{w}_{t}$ to $-\mathbf{w}_{t}$ every $h=256$ steps. The non-stationary behaviour here is that of periodic abrupt flips, similar to the \emph{flipping vector} problem.}

\subsubsection{Stationary bandit problems}
\label{sec:experiments:bandit_problems:stationary}

We further examine the robustness of our proposed approach by evaluating it on selected \emph{stationary} bandit problems.

\paragraph{Stationary Bernoulli.} The mean of the Bernoulli reward for each arm is a constant. We chose $K=8$ arms, with means in $\{ 0.1, 0.2, \ldots, 0.9 \} \setminus \{ 0.5\}$.

\paragraph{Stationary vector.} The observation $x_{t-1}$ for time step $t-1$ encodes a vector $\mathbf{x}_{t-1,k}$ for each action $k$. Each of these vectors is drawn from a finite set of unit vectors in $\mathbb{R}^d$, without replacement within a time step. The probability density function for the reward at time step $t$ given the history $h_{t - 1}$, the action $a_{t}$, and a parameter vector $\mathbf{w}$ is given by $p(r_t \mid h_{t-1}, a_{t}, \mathbf{w}) = \mathcal{N}(r_t \mid \mathbf{w} \cdot \mathbf{x}_{t-1,a_{t}}, s^2)$, where $s^2 > 0$ is a variance. The parameter vector $\mathbf{w}$ is a randomly chosen unit vector. This problem is equivalent to a flipping vector problem where $h \to \infty$. We chose $K=25$ arms, dimension $d=8$, and variance $s^2 = 0.05^2$.

\subsection{Implementation}
\label{sec:implementation}

This section details the implementation of the feedforward and recurrent neural-linear posterior sampling approaches.\footnote{An open-source implementation is available on \url{https://github.com/paulorauber/rnlps}.} Section \ref{sec:evaluation} details the grid search for hyperparameters.%

Feedforward and recurrent neural-linear networks are trained to minimize the mean squared error with an L2 regularization penalty $\lambda = 0.001$.
Every network weight is initially drawn from a standard Gaussian distribution, and redrawn if far from the mean by two standard deviations, and every network bias is initially zero. 
A sequence of $e$ training steps is performed every $q$ time steps (interactions with the environment) using Adam \citep{kingma2014adam} with a learning rate $\eta$. Each training step requires computing the gradient of the loss on the entire dataset or sequence.
The linear regression posterior is recomputed using the entire dataset at every time step. The prior hyperparameters are $\mathbf{w}_0 = \mathbf{0}$ and $\mathbf{V}_{0} = \tau^2 \mathbf{I}$, where $\tau^2 > 0$ is another hyperparameter. One forward pass is required to evaluate each available action. In the recurrent case, note that this does not require forward passing the entire sequence for each action.

\textbf{Feedforward neural-linear posterior sampling.} The feature map $\bs{\psi}$ encodes the last observation $x_{t-1}$ and the action $a_{t}$ together with the last $n$ triplets of observations, actions, and rewards  $\{ (x_{t - k - 1}, a_{t - k}, r_{t-k}) \}_{k = 1}^n$, where $n$ is the so-called order. All actions are one-hot encoded. The feature map $\bs{\psi}$ also encodes the current time step $t$, which is the sole input to a sinusoidal layer with $D$ units. Each sinusoidal unit $i$ computes $\sin(a_i t + b_i)$, where $a_i$ and $b_i$ are network parameters. The output of this sinusoidal layer is concatenated with the remaining inputs from the feature map, comprising the input to the remaining network. This network has three additional hidden layers. The first hidden layer has $L_1$ linear units. The second and third hidden layers have $L_2$ and $L_3$ hyperbolic tangent units, respectively. The last layer has one linear unit (with no bias). 
For the two non-stationary contextual linear bandit problems (see Section \ref{sec:experiments:bandit_problems:contextual}), the feature map $\bs{\psi}$ encodes the action-dependent vector $\mathbf{x}_{t-1,a_{t}}$ instead of any observation $x_{t-1}$, while the corresponding action $a_t$ is not encoded.

\textbf{Recurrent neural-linear posterior sampling.} At a given time step $t$, the input to the recurrent neural network is the reward $r_{t-1}$, the observation $x_{t-1}$, and the action $a_{t}$. This action is one-hot encoded. The network has three hidden layers. The first hidden layer has $L_1$ linear units. The second hidden layer has $L_2$ long short-term memory units \citep{hochreiter1997long, gers2000learning}. The third hidden layer has $L_3$ hyperbolic tangent units. The last layer has one linear unit (with no bias).
For the two non-stationary contextual linear bandit problems, at a given time step $t$, the input to the recurrent neural network is the reward $r_{t-1}$ and the action-dependent vector $\mathbf{x}_{t-1,a_{t}}$, while the corresponding action $a_t$ is not an input.

\subsection{Evaluation}
\label{sec:evaluation}

We present results of at least five policies for each bandit problem. The \emph{random} policy chooses arms at random. The \emph{best (R)NN} policy employs (feedforward or recurrent) neural-linear posterior sampling with hyperparameters that achieve maximum cumulative reward averaged over five independent trials according to an independent grid search for each problem. In contrast, the \emph{default (R)NN} policy employs (feedforward or recurrent) neural-linear posterior sampling with hyperparameters that perform well across either the contextual or the non-contextual problems (including two variations of the rotating vector problem). Concretely, such \emph{default hyperparameters} achieve maximum \emph{normalized score} averaged across either the non-contextual or the contextual problems (from sections \ref{sec:experiments:bandit_problems:non_contextual} and \ref{sec:experiments:bandit_problems:contextual} respectively). The normalized score of a hyperparameter setting $\xi$ on a problem is given by $(m_{\xi} - m_{-})/(m_{+} - m_{-})$, where $m_{\xi}$ is the average cumulative reward of the setting $\xi$ over five independent trials, $m_{-}$ is the average cumulative reward of the random policy, and $m_{+}$ is the average cumulative reward of the corresponding best (R)NN policy. The default hyperparameters allow us to understand whether there are hyperparameter settings that work well for multiple problems. 

Preliminary experiments were employed to choose suitable hyperparameter ranges for the neural-linear approaches. Table \ref{tab:hgrid_defaults} contains a complete description of the hyperparameter grid and the resulting default hyperparameters for each neural-linear approach. More hyperparameter settings are considered for the feedforward approach ($576$) than for the recurrent approach ($96$), which is potentially advantageous for the feedforward approach.

For some of the bandit problems, we also present results of policies based on more conventional algorithms. For non-contextual bandit problems, we present the results of policies based on discounted UCB (D-UCB) and sliding-window UCB (SW-UCB) \citep{garivier2011upper}. The hyperparameters for each of these algorithms were selected based on the same protocol used to select hyperparameters for the best (R)NN policy. %
For D-UCB, the hyperparameter grid contains candidates for the discount factor $\gamma \in \{0.8, 0.85, 0.9, 0.925, 0.95, 0.97, 0.98, 0.99, 0.995, 0.999\}$. For SW-UCB, the hyperparameter grid contains candidates for the window length $\tau \in \{5, 10, 25, 50, 75, 100, 150, 200, 250, 300\}$.

For the non-stationary contextual \emph{linear} bandit problems, we present the results of policies based on discounted linear UCB (D-LinUCB) \citep{russac2019dlinucb} and sliding-window linear UCB (SW-LinUCB) \citep{cheung2019swlinucb}. The hyperparameters for each of these algorithms were selected optimally based on the total number of time steps and the variation budget of each problem \citep{russac2019dlinucb}, requiring additional knowledge in comparison with the neural-linear approaches.

The definitive ten independent trials for each combination of non-contextual bandit problem and policy have double the length of the hyperparameter search trials in order to enable a more conclusive regret analysis. Note that it is quite difficult to establish appropriate trial lengths \emph{before} hyperparameter search.

\begin{table*}[htbp]
    \small
    \centering
    \caption{Hyperparameter grid and default hyperparameters for neural-linear approaches.}
    \label{tab:hgrid_defaults}
    \subfloat{
    \begin{tabular}{l c cc}
      \toprule
      & & \multicolumn{2}{c}{Non-contextual problems} \\
      \cmidrule(lr){3-4}
      Hyperparameter & Candidates & NN & RNN \\
      \midrule
      Learning rate $\eta$ & $\{0.001, 0.01, 0.1\}$ & 0.1 & 0.01 \\
      Number of epochs $e$ by training step & $\{16, 64 \}$ & 16 & 16\\
      Interval $q$ between training steps & $\{32, 128\}$ & 32 & 32\\
      Assumed variance $\sigma^2$ of the reward & $\{0.1, 0.3\}$ & 0.1 & 0.1\\
      Variance $\tau^2$ of the prior distribution & $\{0.5, 1\}$ & 1 & 0.5\\
      Units per layer & $\{ (16,16,16), (32,32,32)\}$ & (32,32,32) & (32,32,32) \\
      Order $n$ & $\{1, 4\}$ & 1 & - \\
      Number of sinusoidal units $D$ & $\{1, 2, 4 \}$ & 1 & -\\
      \bottomrule
    \end{tabular}
    }
    \quad
    \subfloat{
    \begin{tabular}{l c cc}
      \toprule
      & & \multicolumn{2}{c}{Contextual problems} \\
      \cmidrule(lr){3-4}
      Hyperparameter & Candidates & NN & RNN \\
      \midrule
      Learning rate $\eta$ & $\{0.001, 0.01, 0.1\}$ & 0.01 & 0.001 \\
      Number of epochs $e$ by training step & $\{16, 64 \}$ & 64 & 64\\
      Interval $q$ between training steps & $\{32, 128\}$ & 32 & 32\\
      Assumed variance $\sigma^2$ of the reward & $\{0.1, 0.3\}$ & 0.1 & 0.3\\
      Variance $\tau^2$ of the prior distribution & $\{0.5, 1\}$ & 1 & 0.5\\
      Units per layer & $\{ (32,32,32), (64,64,64)\}$ & (32,32,32) & (32,32,32) \\
      Order $n$ & $\{1, 4\}$ & 1 & - \\
      Number of sinusoidal units $D$ & $\{2, 4, 8 \}$ & 2 & -\\
      \bottomrule
    \end{tabular}
    }
\end{table*}
\clearpage

\subsection{Results}
\label{sec:analysis}

Section \ref{app:sec:results:regret_non_contextual} and Section \ref{app:sec:results:regret_contextual_problems} present a regret curve for each combination of problem and policy. Each of these curves aggregates the empirical regret across ten independent trials (not considered for hyperparameter search), and shows bootstrapped confidence intervals of $95\%$. The average empirical regret at the end of these trials is summarized in Table \ref{tab:average_regrets_w_stddev_main}.

Section \ref{app:sec:results:sensitivity_non_contextual} and Section \ref{app:sec:results:sensitivity_contextual} present a hyperparameter sensitivity curve for each combination of problem and neural-linear approach. A hyperparameter sensitivity curve displays the average cumulative reward achieved by each hyperparameter setting (sorted from highest to lowest along the horizontal axis). Such curves are useful to assess robustness regarding hyperparameter choices.

These results are analysed in Section \ref{app:sec:results:analysis}.

\newpage

\subsubsection{Regret curves: non-contextual problems}
\label{app:sec:results:regret_non_contextual}

\begin{figure}[h!]
    \centering
    \begin{floatrow}
      \ffigbox{\caption{Flipping Gaussian.} \label{fig:flipping_gaussian}}{%
        \includegraphics[width=0.9\linewidth]{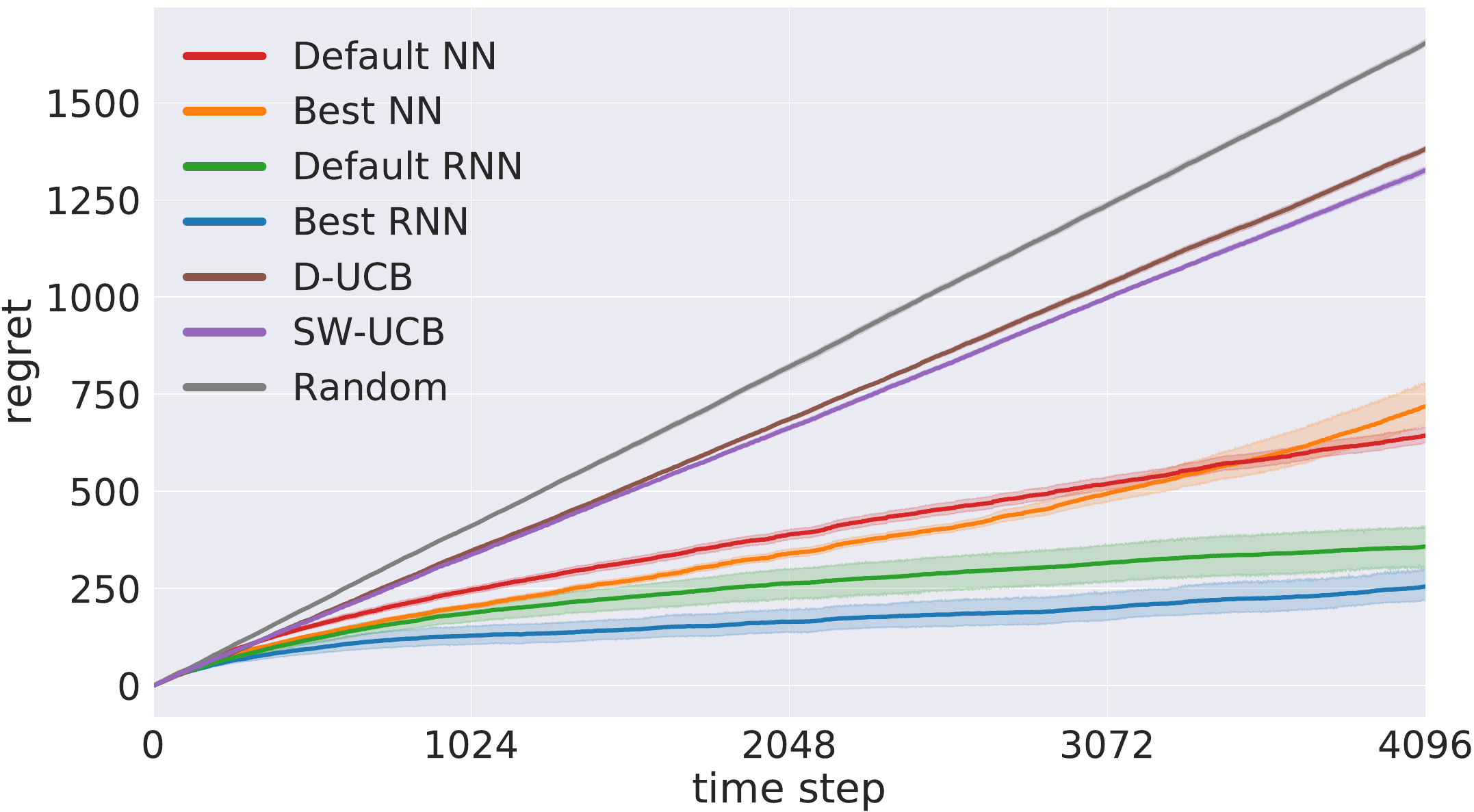}
      }
      \ffigbox{\caption{Flipping Bernoulli.} \label{fig:flipping_bernoulli}}{%
        \includegraphics[width=0.9\linewidth]{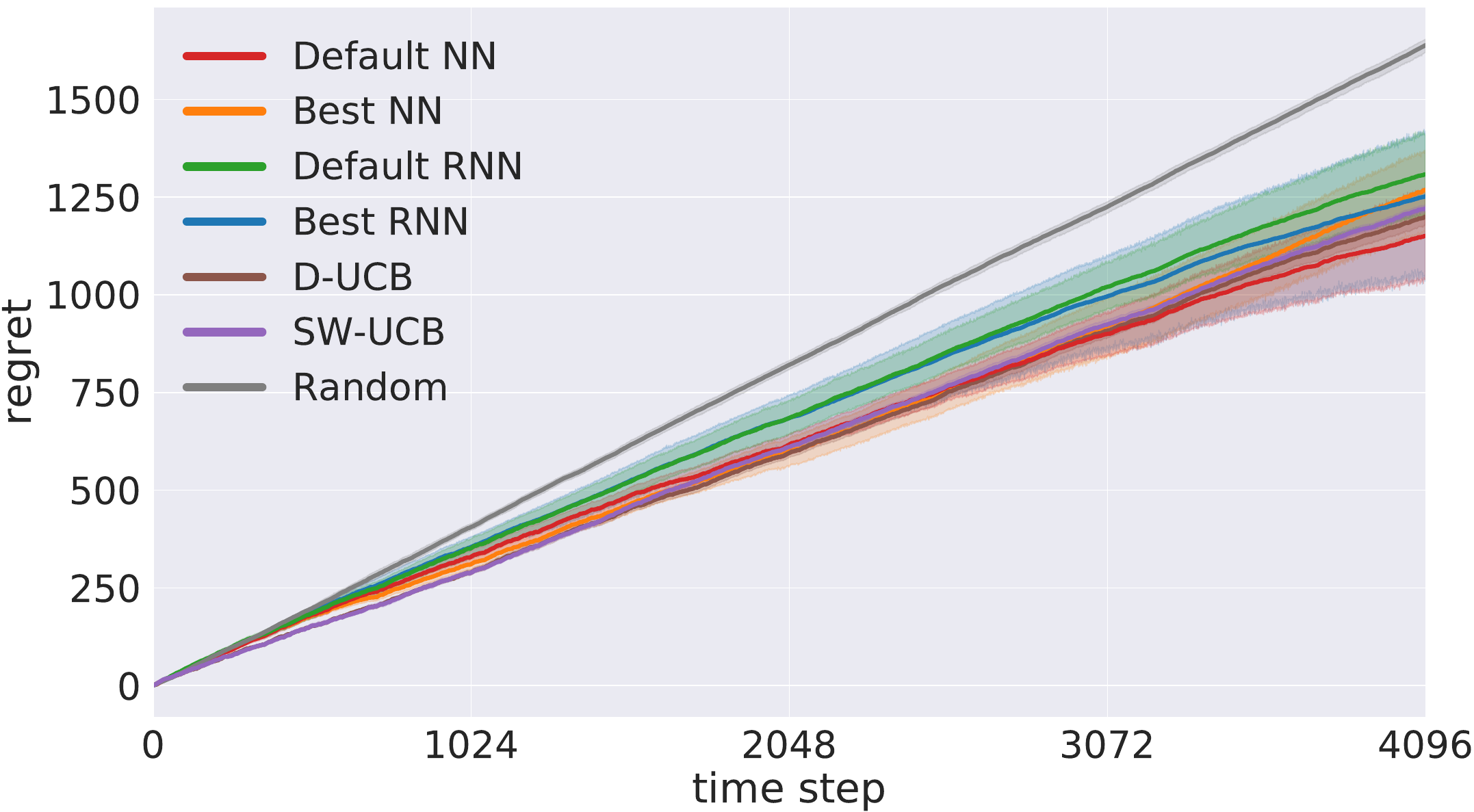}
      }
    \end{floatrow}
    \vspace{0.7cm}
    \begin{floatrow}
      \ffigbox{\caption{Sinusoidal Bernoulli.}\label{fig:sinusoidal_bernoulli}}{%
        \includegraphics[width=0.9\linewidth]{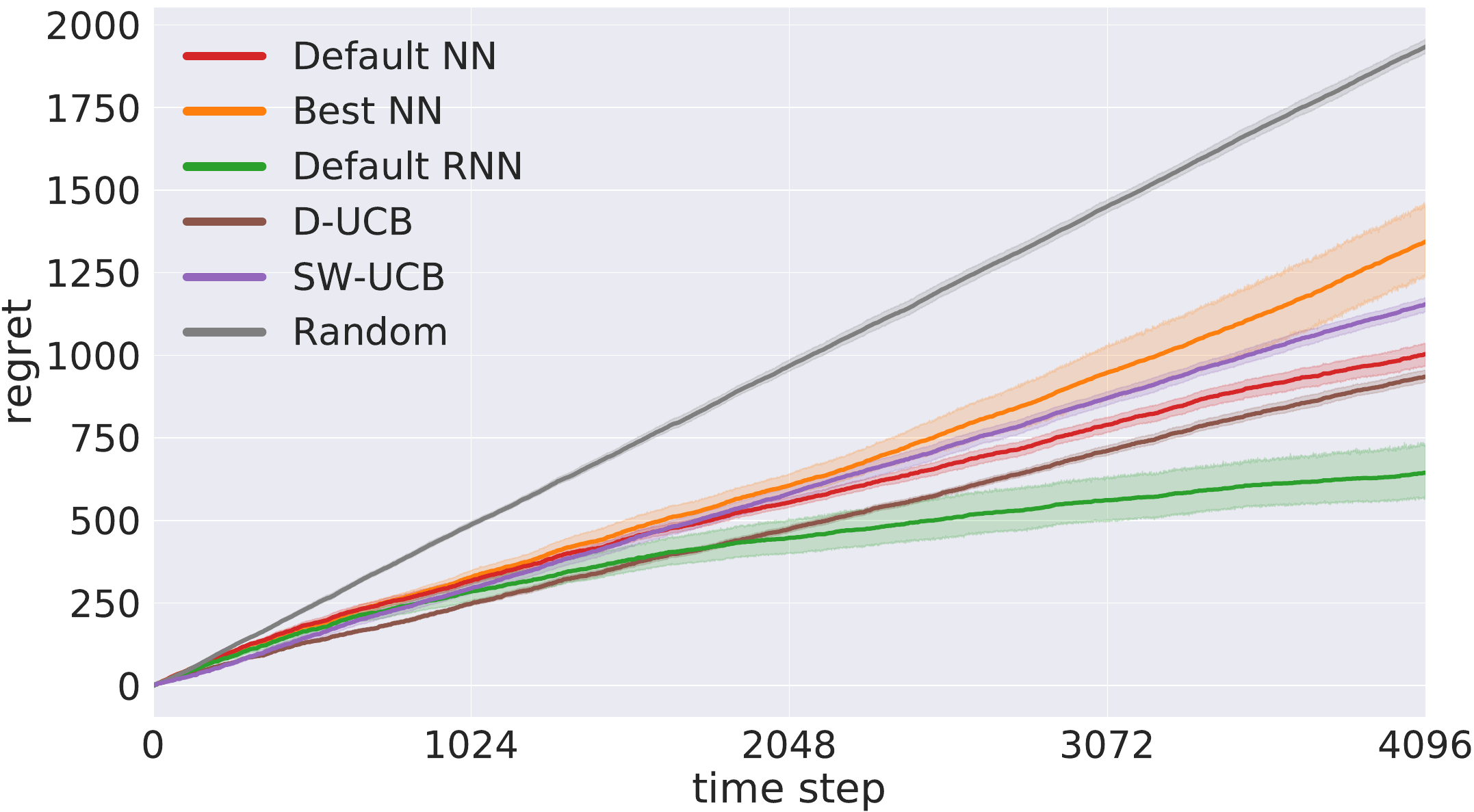}
      }
      \ffigbox{\caption{Circular Markov chain.}\label{fig:circular_markov_chain}}{%
        \includegraphics[width=0.9\linewidth]{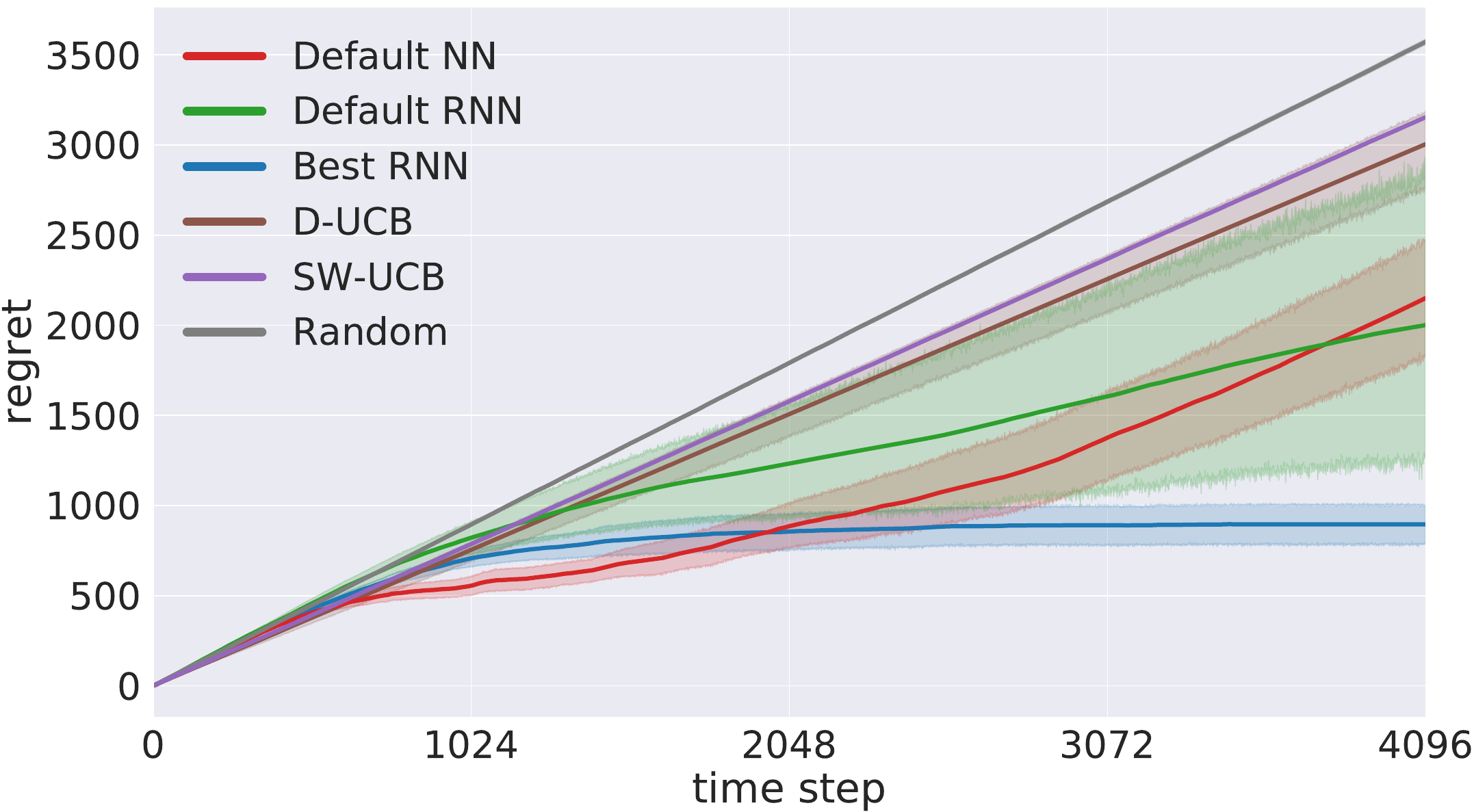}
      }
    \end{floatrow}
\end{figure}

\newpage
\subsubsection{Regret curves: contextual problems}
\label{app:sec:results:regret_contextual_problems}

\begin{figure}[h!]
    \centering
    \begin{floatrow}
      \ffigbox{\caption{Flipping digits.} \label{fig:flipping_digits}}{%
        \includegraphics[width=1.0\linewidth]{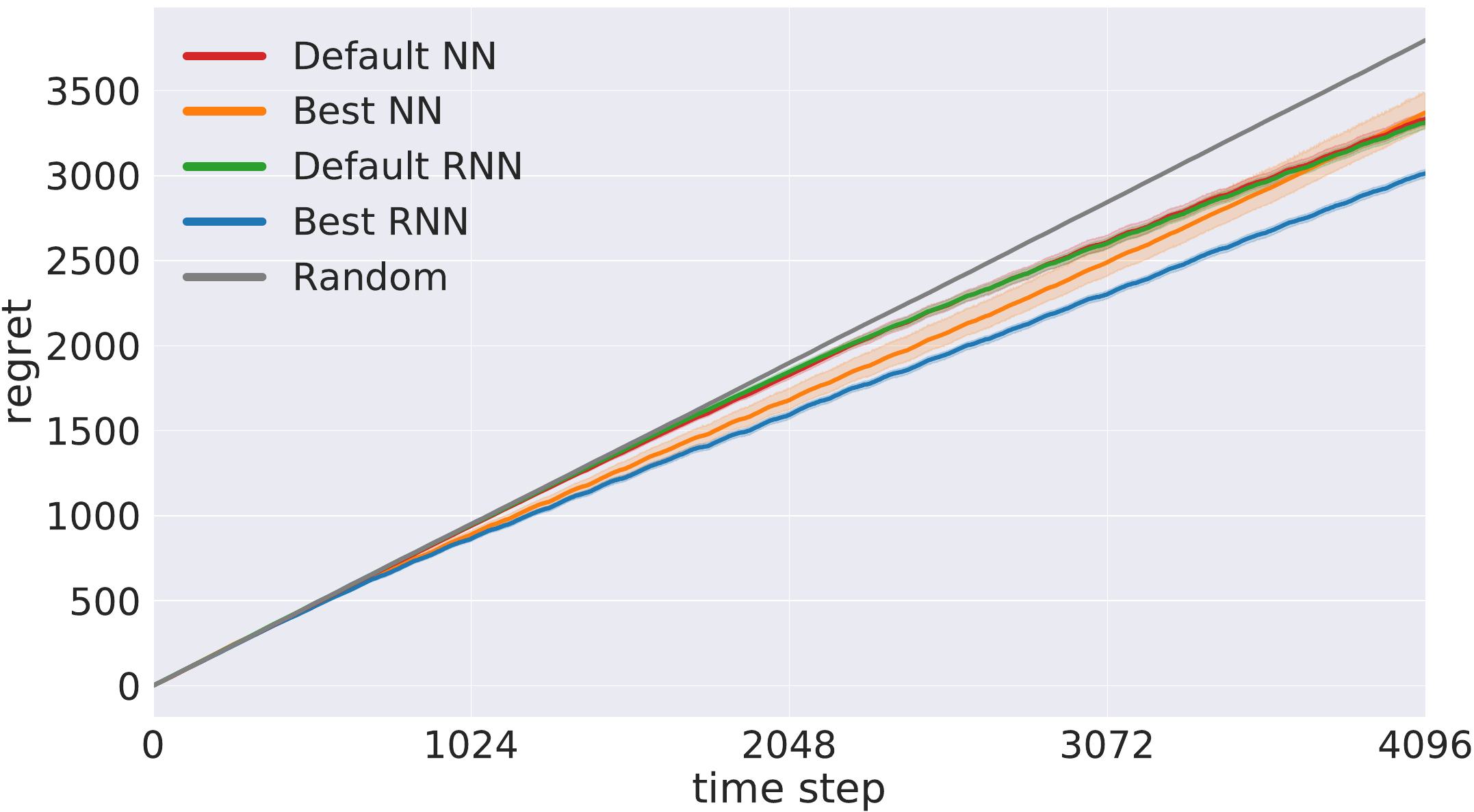}
      }
      \ffigbox{\caption{Wall-following robot.} \label{fig:wall_following}}{%
        \includegraphics[width=1.0\linewidth]{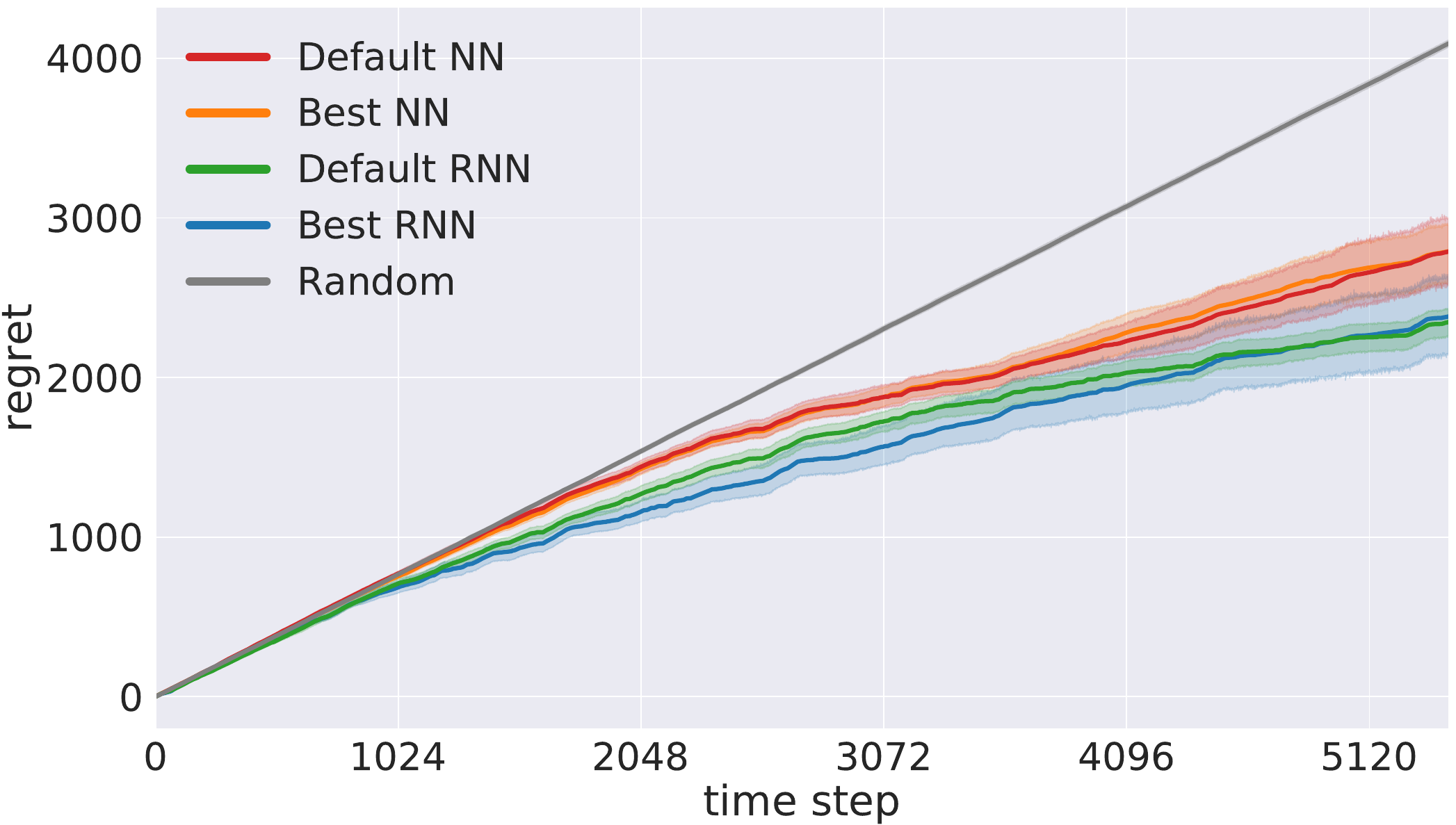}
      }
    \end{floatrow}
    \vspace{0.7cm}
    \begin{floatrow}
      \ffigbox{\caption{Flipping vector.} \label{fig:flipping_vector}}{%
        \includegraphics[width=1.0\linewidth]{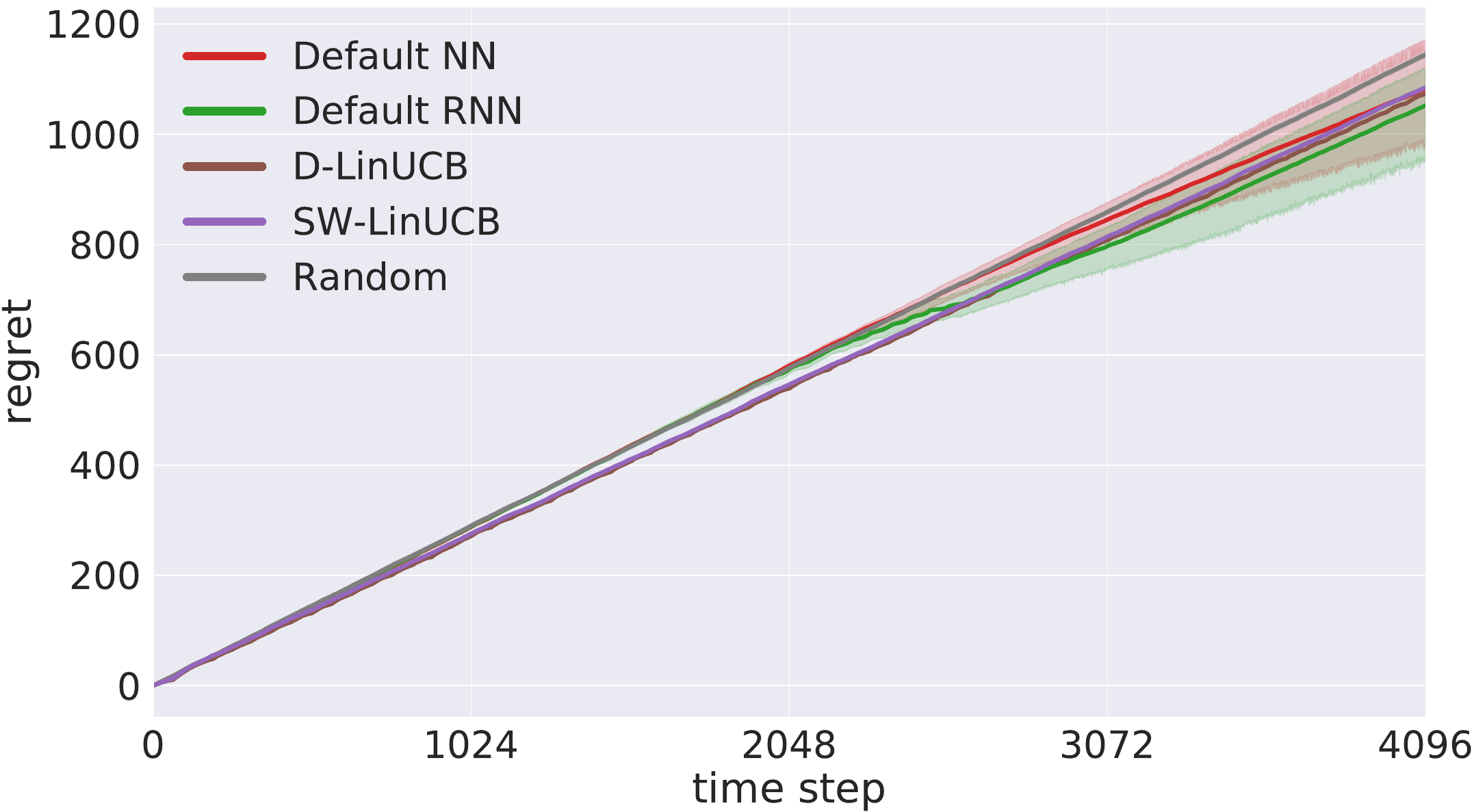}
      }
      \ffigbox{\caption{Rotating vector ($f = 32^{-1}$).}\label{fig:rotating_vector_32}}{%
        \includegraphics[width=1.0\linewidth]{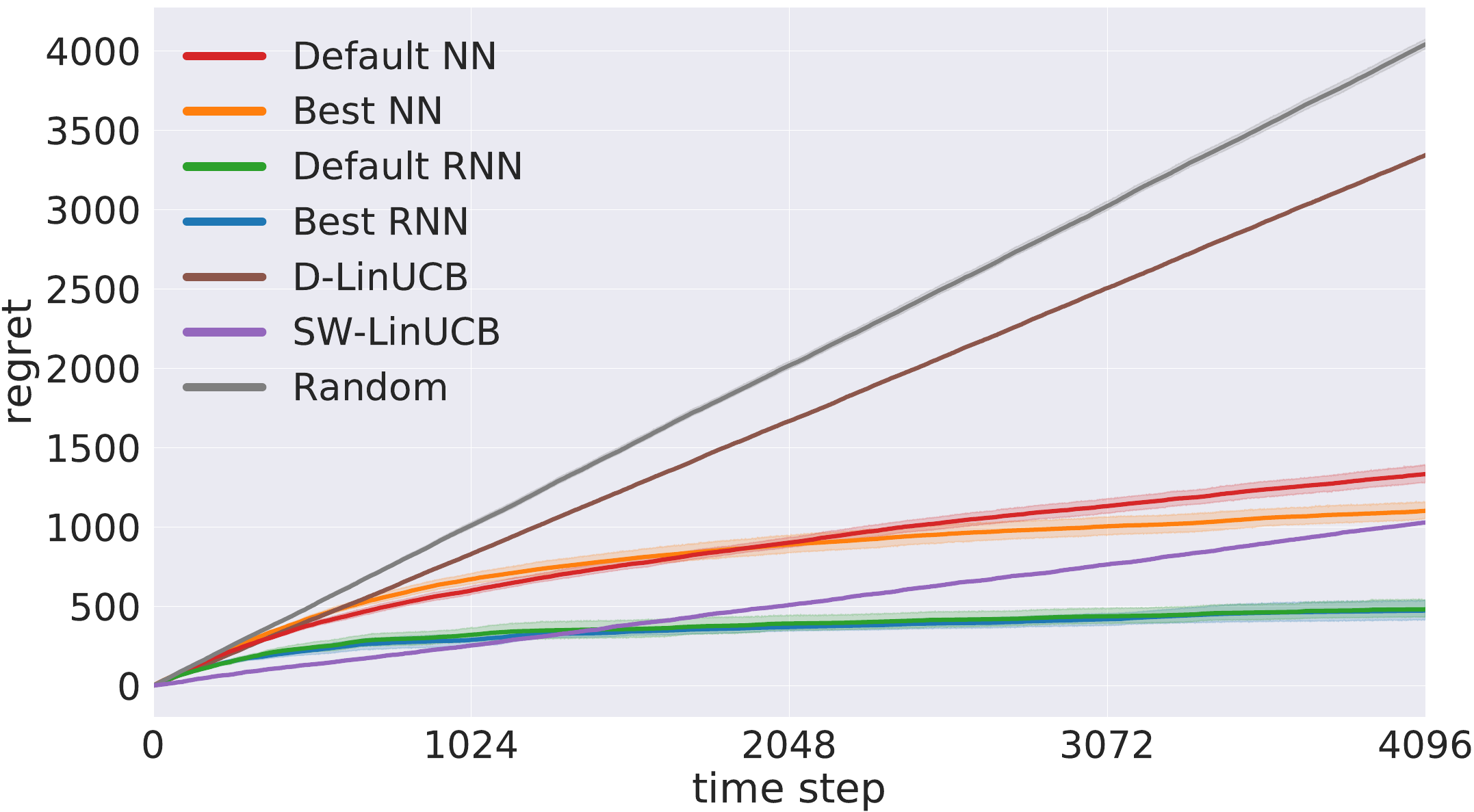}
      }
    \end{floatrow}
    \vspace{0.7cm}
    \begin{floatrow}
      \ffigbox{\caption{Rotating vector ($f = 2048^{-1}$).}\label{fig:rotating_vector_2048}}{%
        \includegraphics[width=1.0\linewidth]{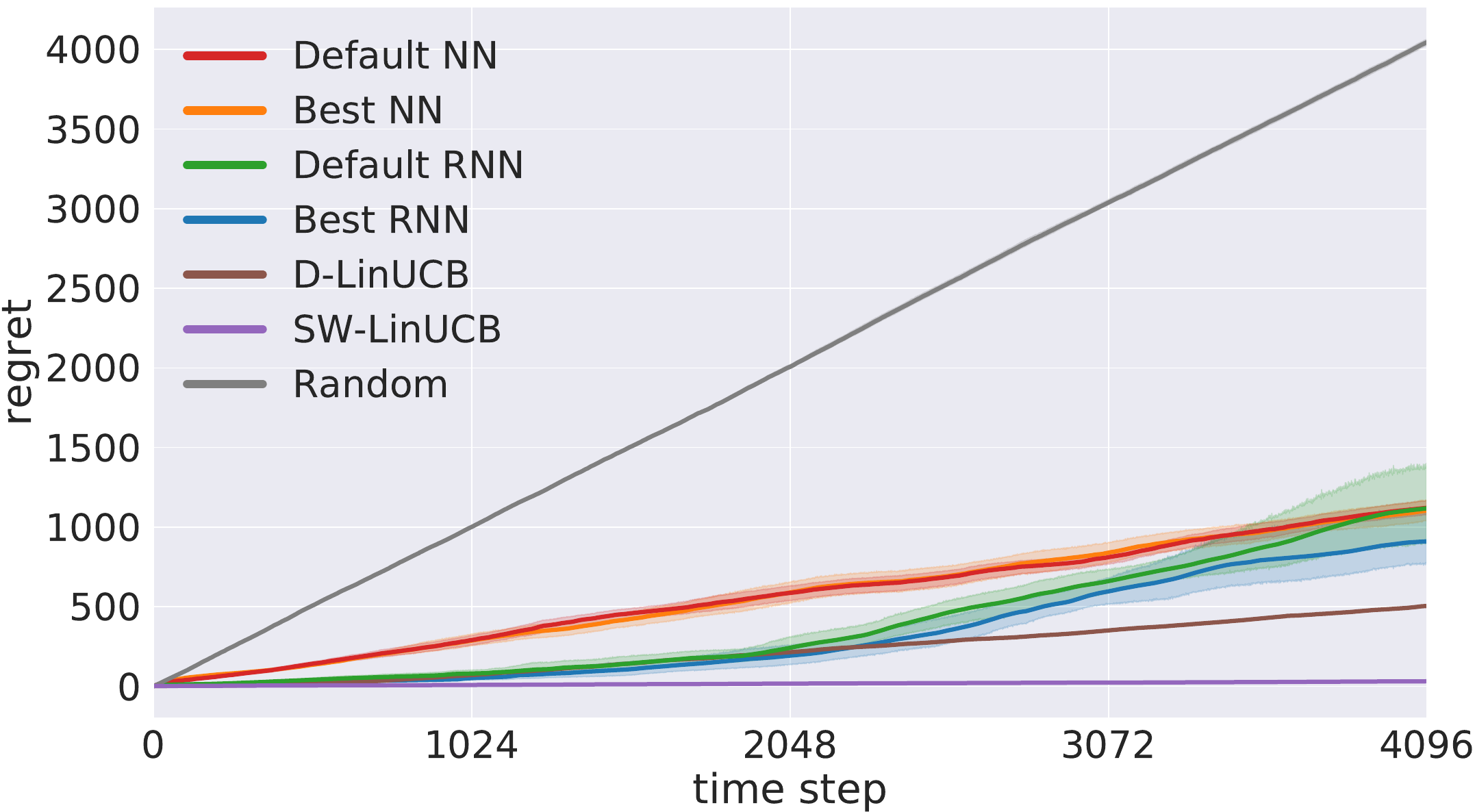}
      }
    \end{floatrow}
\end{figure}

\newpage

\subsubsection{Hyperparameter sensitivity plots: non-contextual problems}
\label{app:sec:results:sensitivity_non_contextual}
\begin{figure}[h!]
    \centering
    \begin{floatrow}
      \ffigbox{\caption{Flipping Gaussian.}}{%
        \includegraphics[width=1.0\linewidth]{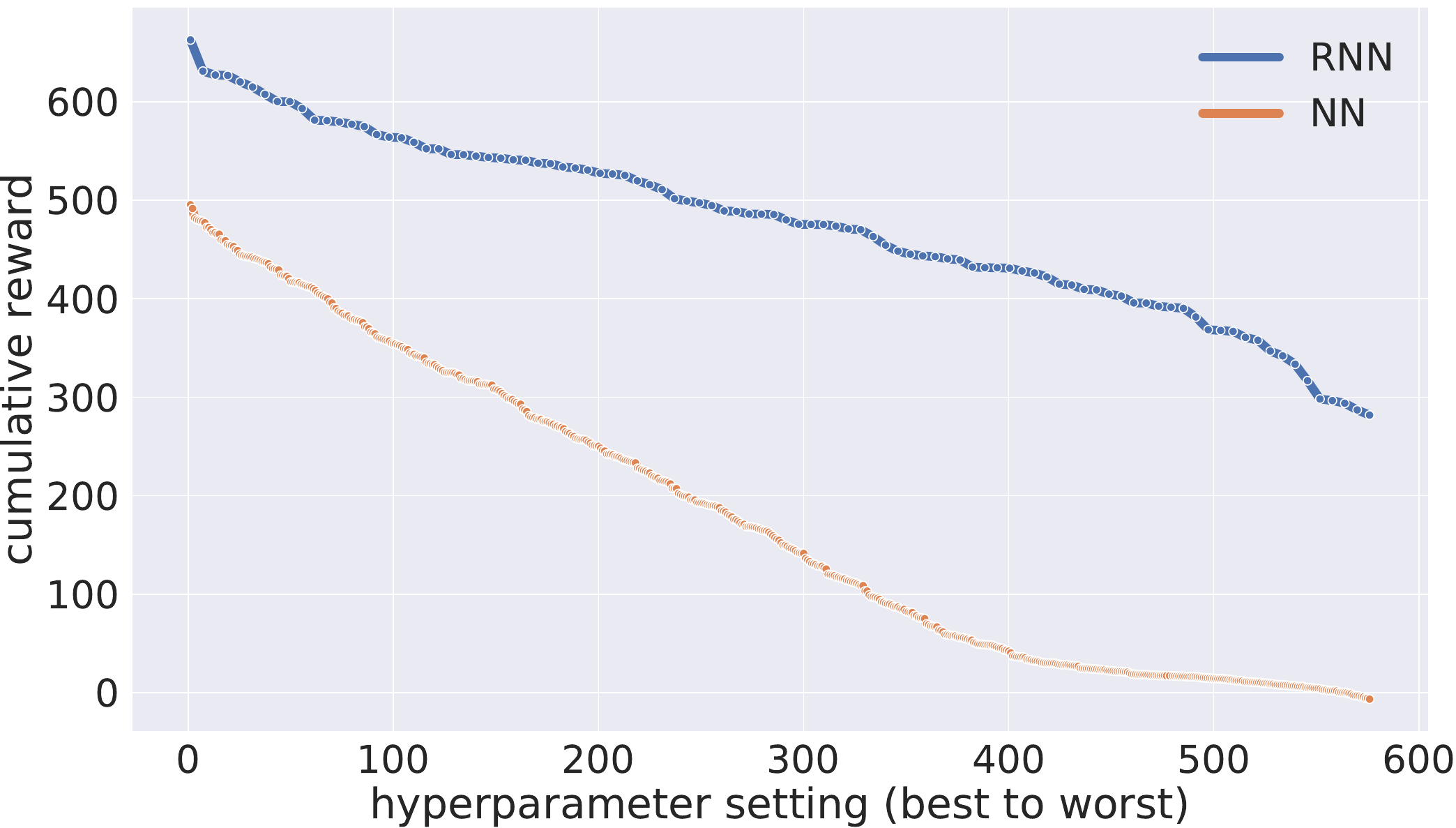}
      }
      \ffigbox{\caption{Flipping Bernoulli.}}{%
        \includegraphics[width=1.0\linewidth]{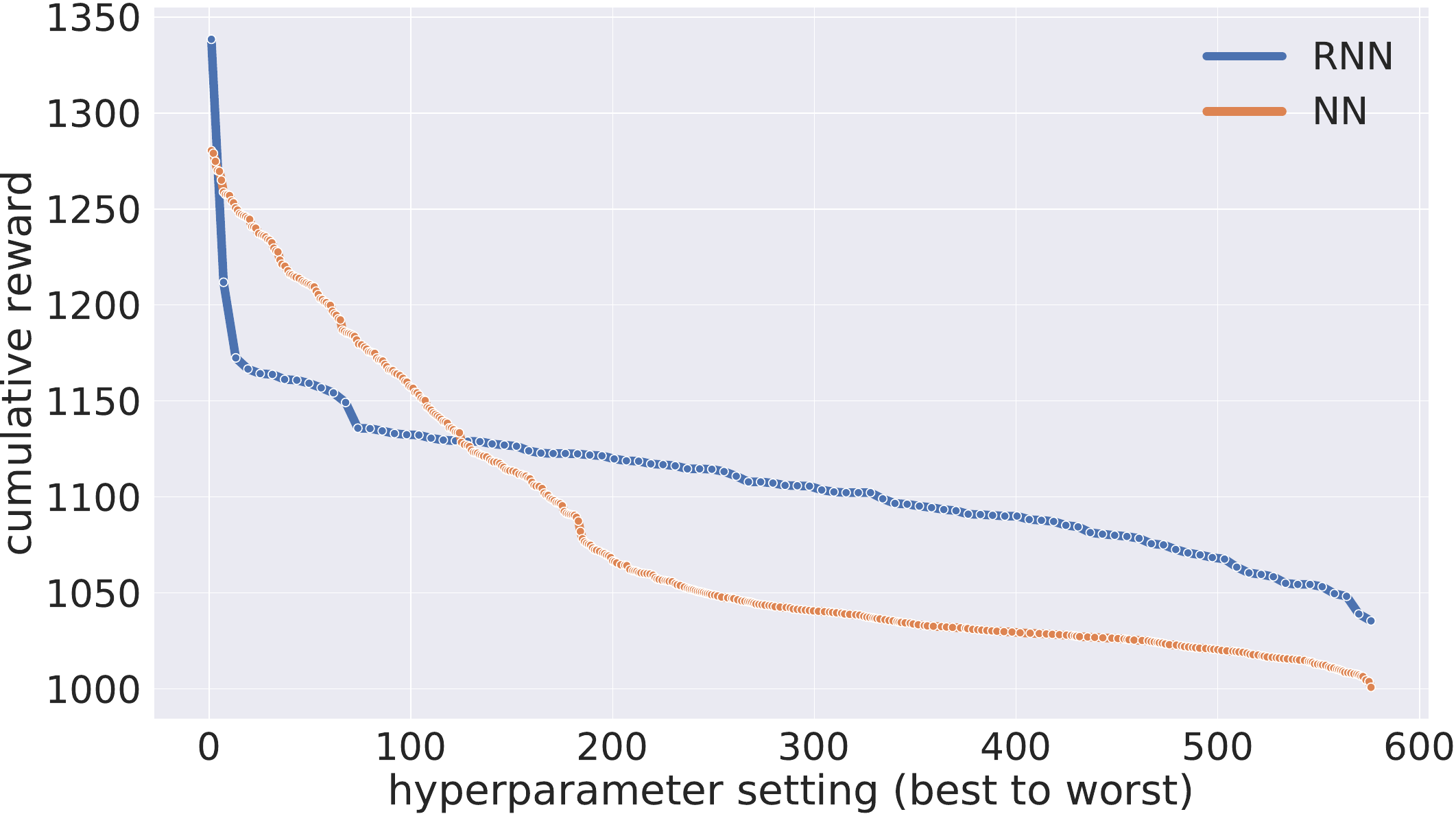}
      }
    \end{floatrow}
    \vspace{0.7cm}
    \begin{floatrow}
      \ffigbox{\caption{Sinusoidal Bernoulli.}}{%
        \includegraphics[width=1.0\linewidth]{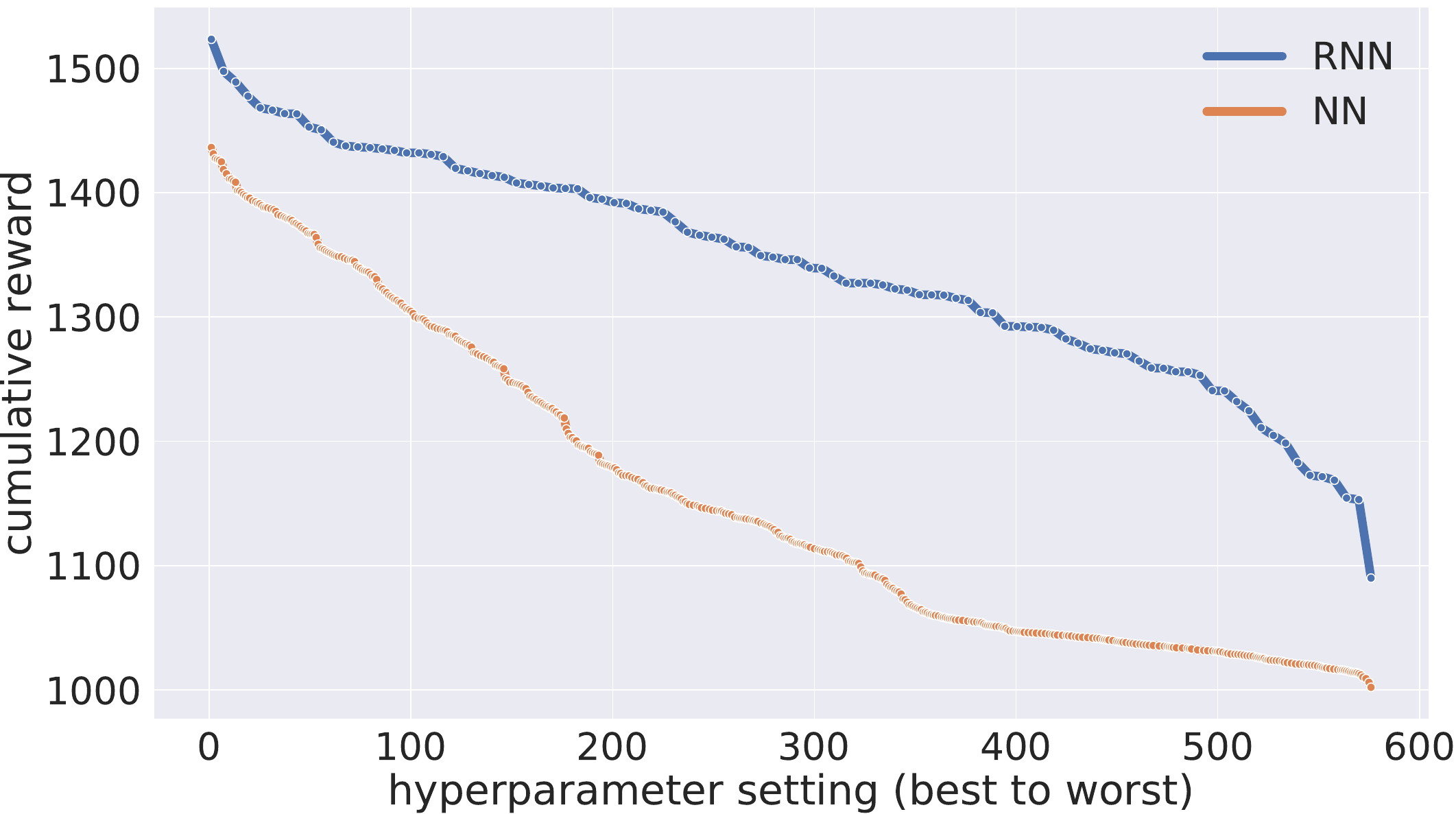}
      }
      \ffigbox{\caption{Circular Markov chain.}}{%
        \includegraphics[width=1.0\linewidth]{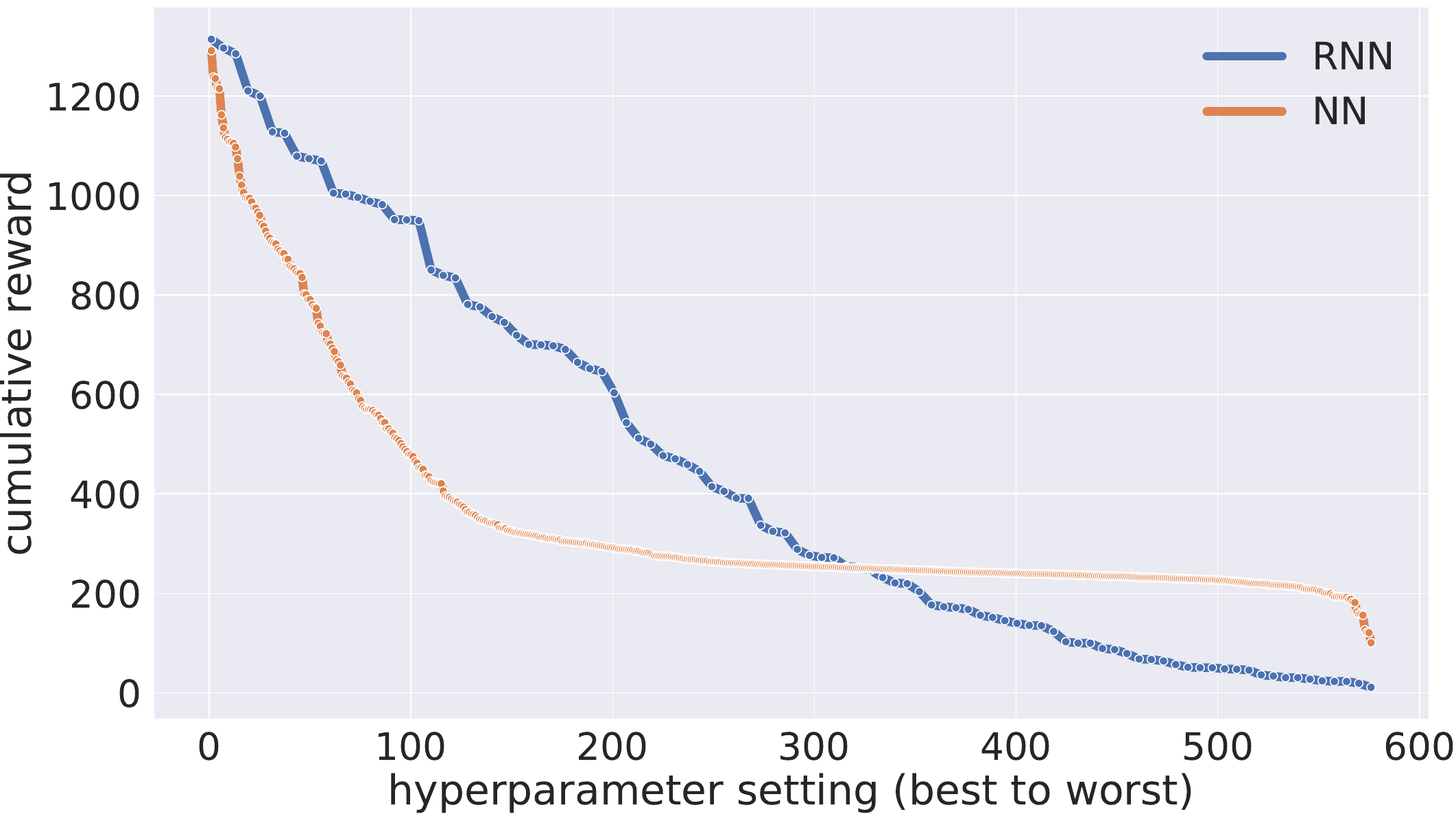}
      }
    \end{floatrow}
\end{figure}

\newpage
\subsubsection{Hyperparameter sensitivity plots: contextual problems}
\label{app:sec:results:sensitivity_contextual}
\begin{figure}[h!]
    \centering
    \begin{floatrow}
      \ffigbox{\caption{Flipping digits.}}{%
        \includegraphics[width=1.0\linewidth]{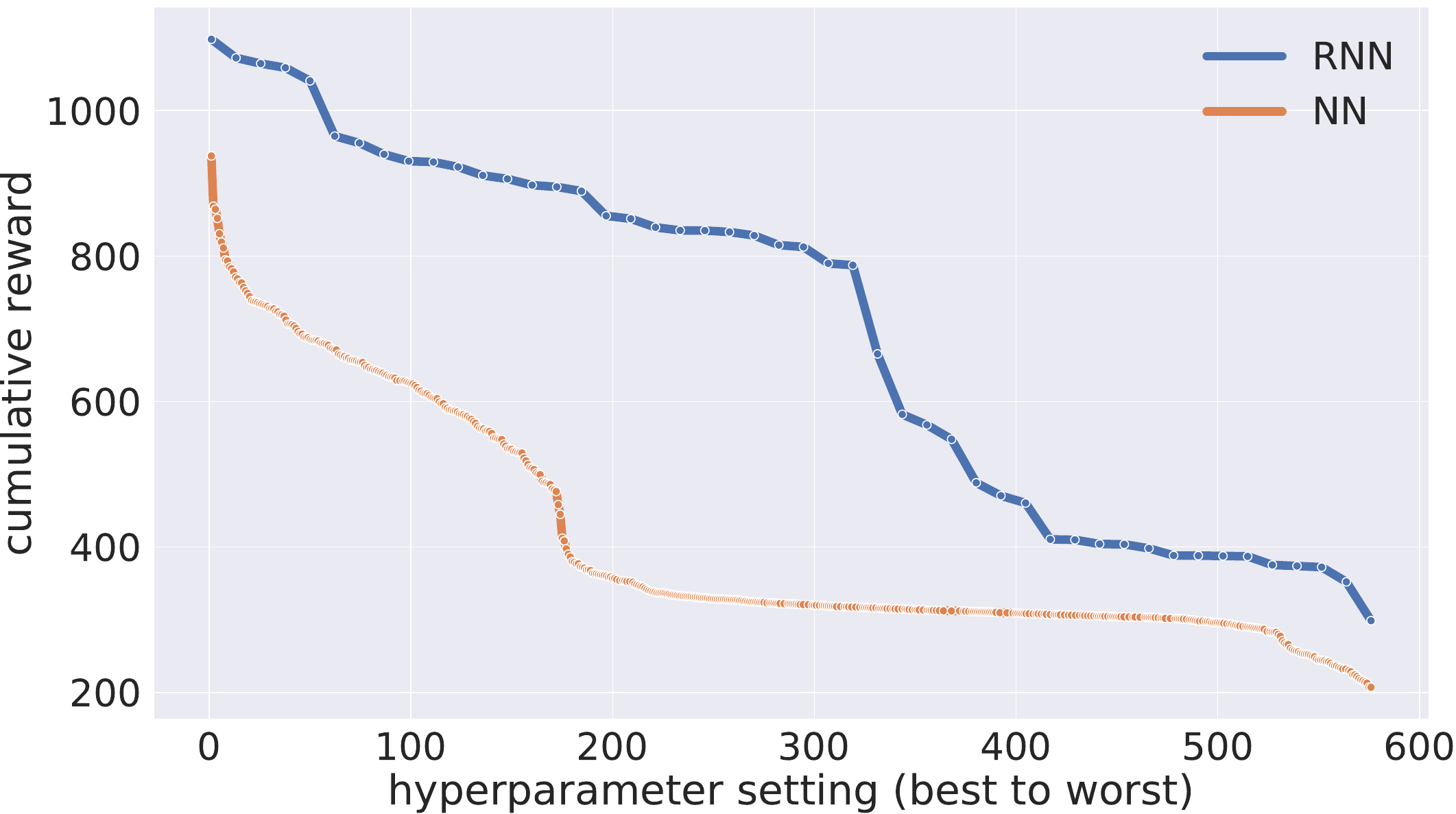}
      }
      \ffigbox{\caption{Wall-following robot.}}{%
        \includegraphics[width=1.0\linewidth]{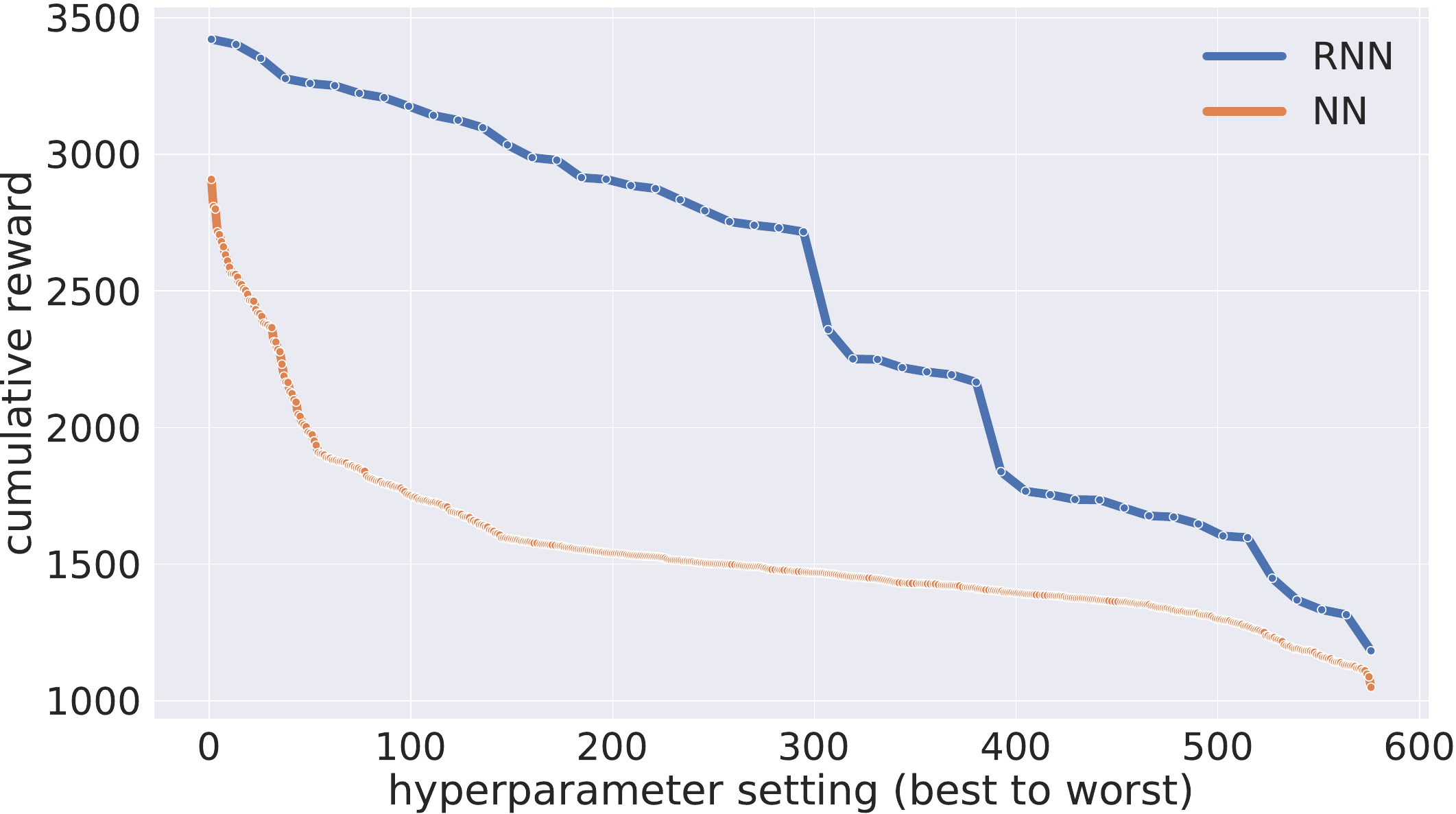}
      }
    \end{floatrow}
    \vspace{0.7cm}
    \begin{floatrow}
      \ffigbox{\caption{Flipping vector.}}{%
        \includegraphics[width=1.0\linewidth]{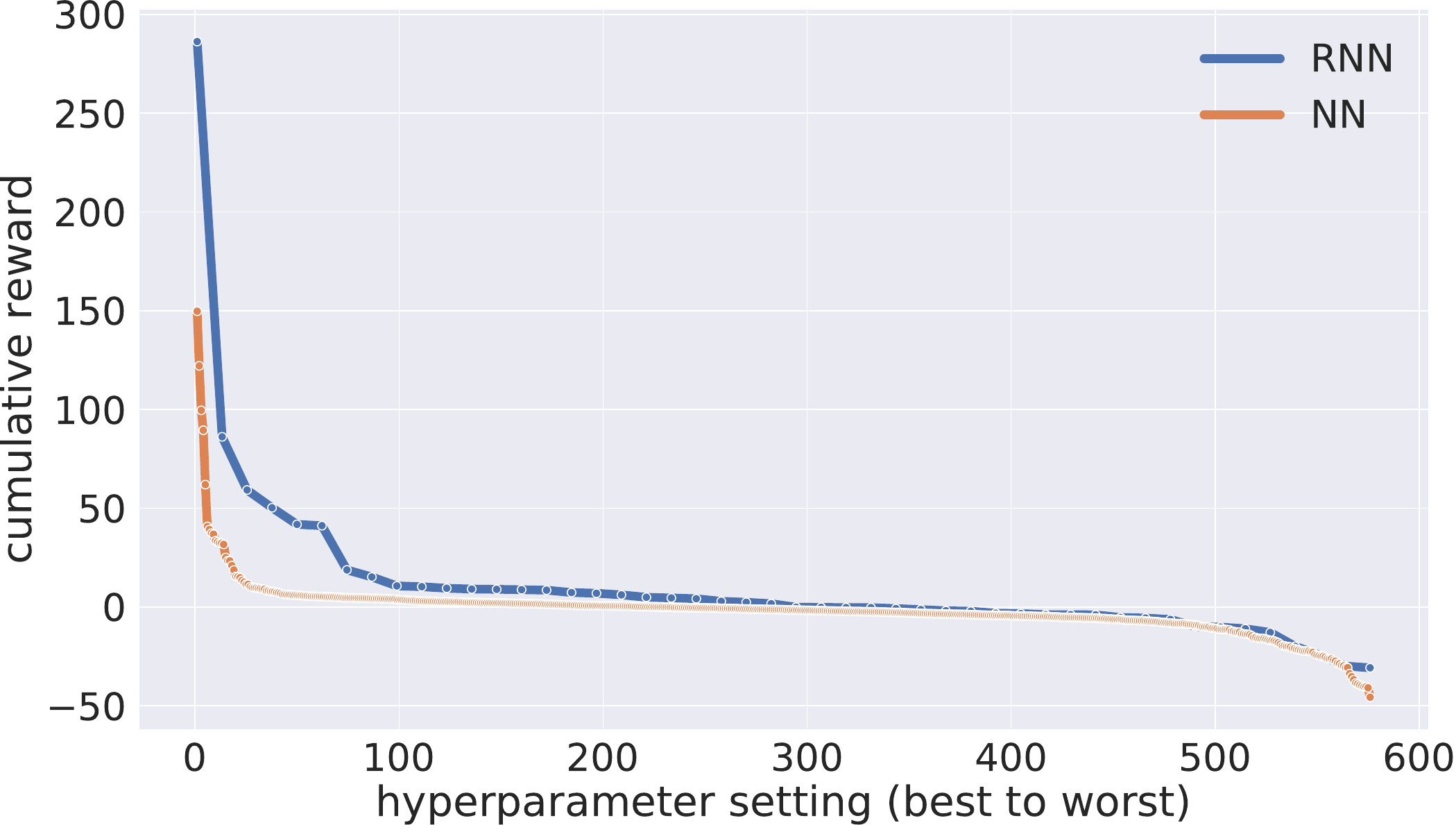}
      }
      \ffigbox{\caption{Rotating vector ($f = 32^{-1}$).}}{%
        \includegraphics[width=1.0\linewidth]{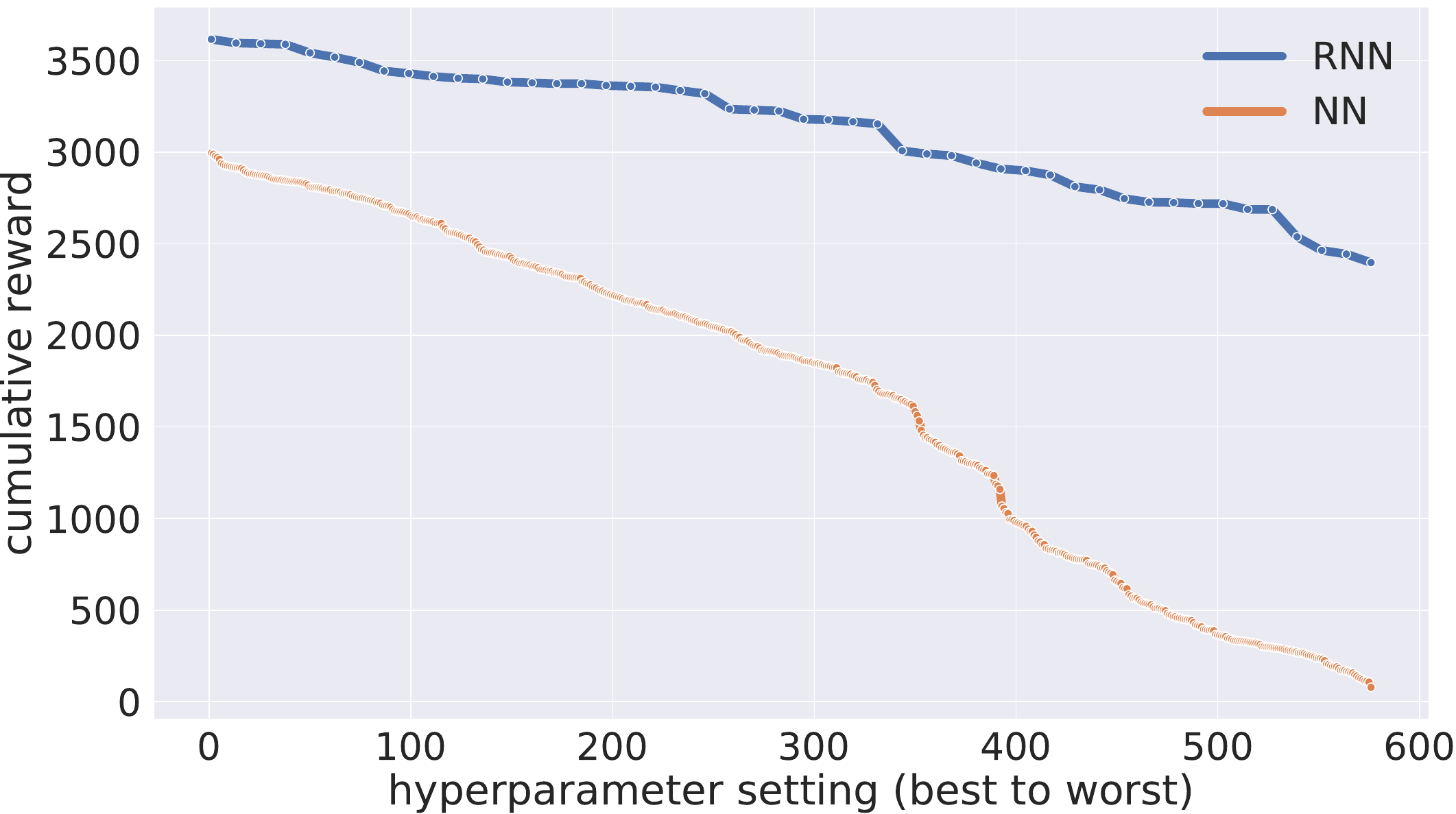}
      }
    \end{floatrow}
    \vspace{0.7cm}
    \begin{floatrow}
      \ffigbox{\caption{Rotating vector ($f = 2048^{-1}$).}}{%
        \includegraphics[width=1.0\linewidth]{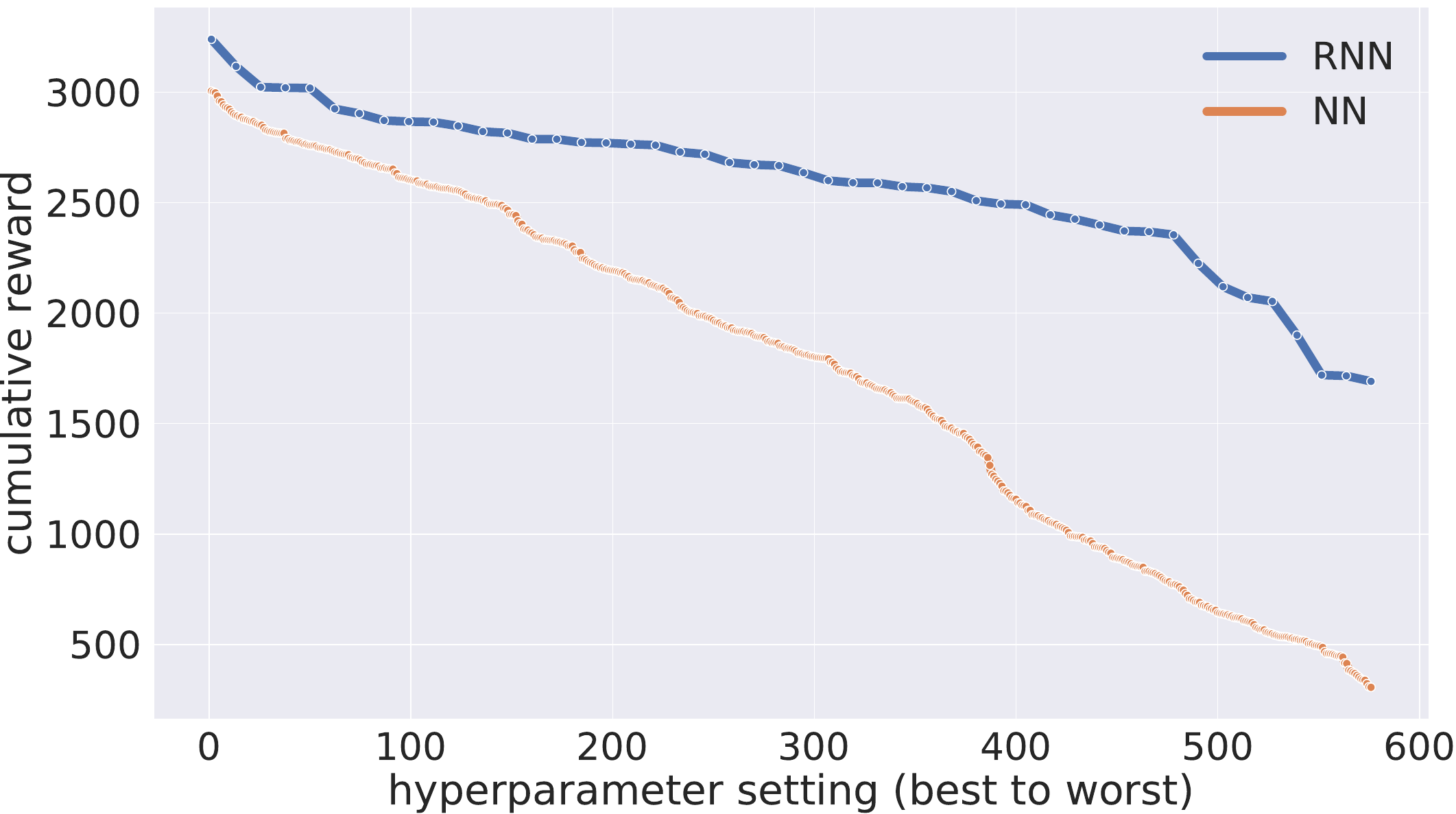}
      }
    \end{floatrow}
\end{figure}

\newpage

\begin{sidewaystable}[th]
\small
    \centering
    \begin{tabular}{c cccc cc}
    \toprule
    Bandit problem & Best NN & Best RNN & Default NN & Default RNN & D-(L)UCB & SW-(L)UCB \\
    \midrule
    Flipping Gaussian & $719.35 \pm 93.54$ & $\bf{254.59} \pm 68.43$ & $643.38 \pm 34.5$ & $357.58 \pm 88.46$ & $1381.3 \pm 13.32$ & $1327.27 \pm 15.55$  \\
    Flipping Bernoulli & $1267.9 \pm 180.28$ & $1251.9 \pm 313.97$ & $\bf{1151} \pm 196.4$ & $1308.5 \pm 180.01$ & $1199.9 \pm 34.91$ & $1220.6 \pm 16.19$  \\
    Sinusoidal Bernoulli & $1344.94 \pm 182.95$ & $\bf{643.94} \pm 138.29$ & $1003.24 \pm 58.46$ & $\bf{643.94} \pm 138.29$ & $935.24 \pm 30.61$ & $1154.64 \pm 36.94$  \\
    C. Markov chain & $2151.97 \pm 758.29$ & $\bf{895.23} \pm 188.75$ & $2151.97 \pm 758.29$ & $2001.57 \pm 1360.54$ & $3005.1 \pm 372.93$ & $3154.96 \pm 16.97$  \\
    \multicolumn{7}{c}{} \\
    Flipping digits & $3372.38 \pm 182.87$ & $\bf{3014.88} \pm 41.88$ & $3334.48 \pm 66.88$ & $3314.28 \pm 64.5$ & - & -  \\
    Wall-following robot & $2791.47 \pm 311.13$ & $2383.47 \pm 420.15$ & $2790.07 \pm 360.8$ & $\bf{2348.27} \pm 147.86$ & - & -  \\
    Flipping vector & $1082.86 \pm 155.53$ & $\bf{1052.22} \pm 146.74$ & $1082.86 \pm 155.53$ & $\bf{1052.22} \pm 146.74$ & $1073.92 \pm 4.7$ & $1084.6 \pm 7.74$  \\
    R. vector ($f = 32^{-1}$) & $1102.34 \pm 89.45$ & $\bf{473.72} \pm 103.27$ & $1333.47 \pm 93.68$ & $482.05 \pm 95.64$ & $3345.18 \pm 4.82$ & $1028.9 \pm 1.49$ \\
    R. vector ($f = 2048^{-1}$) & $1103.81 \pm 110.36$ & $910.81 \pm 297.53$ & $1121.37 \pm 74.59$ & $1118.18 \pm 413.99$ & $504.77 \pm 2.03$ & $\bf{31.52} \pm 1$  \\
    \bottomrule
    \end{tabular}
    \caption{Results across ten trials for each combination of problem and policy.}
    \label{tab:average_regrets_w_stddev_main}
\end{sidewaystable}
\clearpage
\subsubsection{Analysis}
\label{app:sec:results:analysis}
This section highlights the most notable aspects of the results.

\textbf{Non-contextual bandit problems.} 
In the flipping Gaussian problem (Fig. \ref{fig:flipping_gaussian}), the recurrent policies outperform the other policies by a large margin, and their regret grows very slowly by the end of the trials. The conventional non-stationary policies perform very poorly, which illustrates the importance of the fact that the recurrent approach is able to \emph{predict} rather than \emph{react} in order to exploit periodicity. The hyperparameter sensitivity plot  shows that the recurrent approach is also much more robust to hyperparameter choices than the feedforward approach (Sec. \ref{app:sec:results:sensitivity_non_contextual}). 

In the flipping Bernoulli problem (Fig. \ref{fig:flipping_bernoulli}), the default NN outperforms the other non-random policies by an insignificant margin. This problem is much more difficult than the flipping Gaussian problem, as evidenced by the regret that grows quickly for every policy by the end of the trials. The hyperparameter sensitivity plot shows that the feedforward approach is arguably more robust to hyperparameter choices (Sec. \ref{app:sec:results:sensitivity_non_contextual}). Although the best hyperparameters for the recurrent approach outperform the best hyperparameters for the feedforward approach during hyperparameter search, the (longer and more numerous) definitive trials lead to the opposite conclusion. %

In the sinusoidal Bernoulli problem (Fig. \ref{fig:sinusoidal_bernoulli}), the recurrent policies outperform every other policy by a significant margin, and their regret grows slowly by the end of the trials. D-UCB outperforms the default NN policy, which in turn outperforms the best NN policy. The hyperparameter sensitivity plot also heavily favors the recurrent approach (Sec. \ref{app:sec:results:sensitivity_non_contextual}).

In the circular Markov chain problem (Fig. \ref{fig:circular_markov_chain}), the best RNN policy outperforms the remaining policies by a significant margin, and its regret grows slowly by the end of the trials. However, the default RNN policy exhibits an atypical large variance in regret. Because the hyperparameter sensitivity plot does not suggest a lack of robustness for the recurrent approach (Sec. \ref{app:sec:results:sensitivity_non_contextual}), we decided to investigate the cause of this variance, and noticed that the default RNN achieves worst than random performance across three of the ten trials. This suggests that the recurrent approach may benefit from a more careful initialization of recurrent neural network parameters. Unsurprisingly, the conventional non-stationary policies are not able to exploit the structure of this problem.

\textbf{Contextual bandit problems.} In the flipping digits problem (Fig. \ref{fig:flipping_digits}), the best RNN policy outperforms the other non-random policies, which achieve equivalent performance. This is a difficult problem, as evidenced by the regret that grows quickly for every policy by the end of the trials.

In the wall-following robot problem (Fig. \ref{fig:wall_following}), the recurrent policies outperform the feedforward policies by a very small margin, and their regret grows slowly by the end of the dataset. In the flipping vector problem (Fig. \ref{fig:flipping_vector}), the combination of non-stationarity with high-dimensional observations proves too challenging for all policies.

In the low-frequency rotating vector problem ($f = 1/2048$, $2$ rotations per trial, Fig. \ref{fig:rotating_vector_2048}), the conventional non-stationary policies outperform every other policy. This is not surprising, since the corresponding algorithms have access to additional knowledge, and were designed specially for similar problems. More interestingly, in the high-frequency rotating vector problem ($f = 1/32$, $120$ rotations per trial, Fig. \ref{fig:rotating_vector_32}), the recurrent policies significantly outperform every other policy, which once again illustrates the importance of prediction over reaction in order to exploit periodicity. The success of conventional non-stationary policies is highly dependent on the so-called variation budget \citep{russac2019dlinucb}, which explains their poor performance in environments that change quickly.

The hyperparameter sensitivity plots for contextual problems show that the recurrent approach is consistently more robust to hyperparameter choices than the feedforward approach (Sec. \ref{app:sec:results:sensitivity_contextual}).

\subsection{Additional experiments}
\textbf{Experiments using the Criteo dataset.} In this set of experiments, we evaluate the default neural policies on bandit problems created by using a sample of 30 days of the Criteo live traffic data \citep{diemert2017attribution}. The dataset consists of banners that were displayed to users, along with contextual information, and a target variable indicating whether the banner was clicked by the user. We follow the experimental setup used by \citet{kim2020randomized}. Further details are provided in Section \ref{sec:experiments:bandit_problems:criteo}.

\begin{figure}[htb]
\begin{subfigure}{.49\textwidth}
  \centering
  \includegraphics[width=0.99\textwidth]{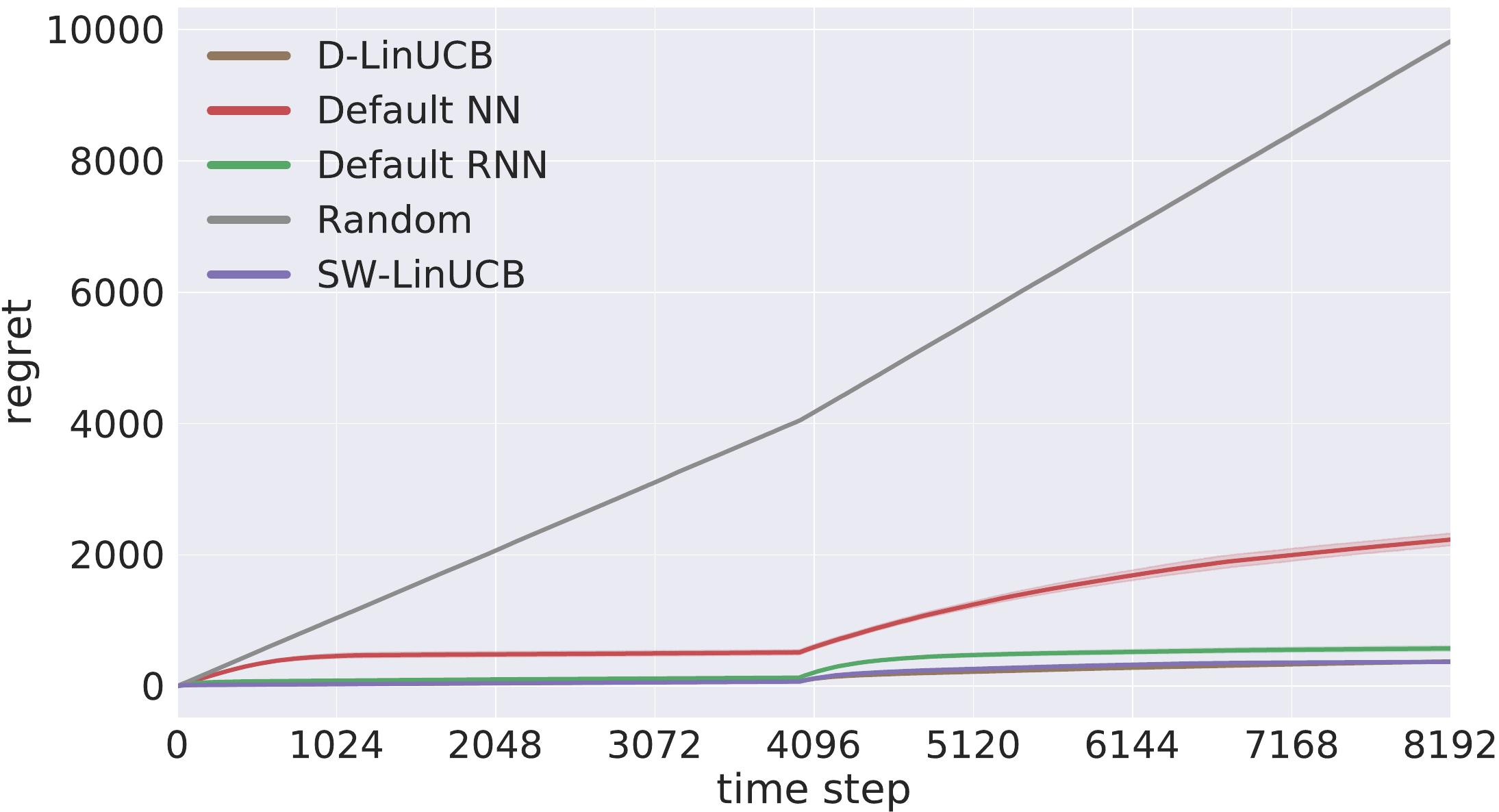}
  \caption{Abrupt partial flip at $t=4000$}
  \label{fig:criteo_exps:abrupt_50}
\end{subfigure}
\begin{subfigure}{.49\textwidth}
  \centering
  \includegraphics[width=0.99\textwidth]{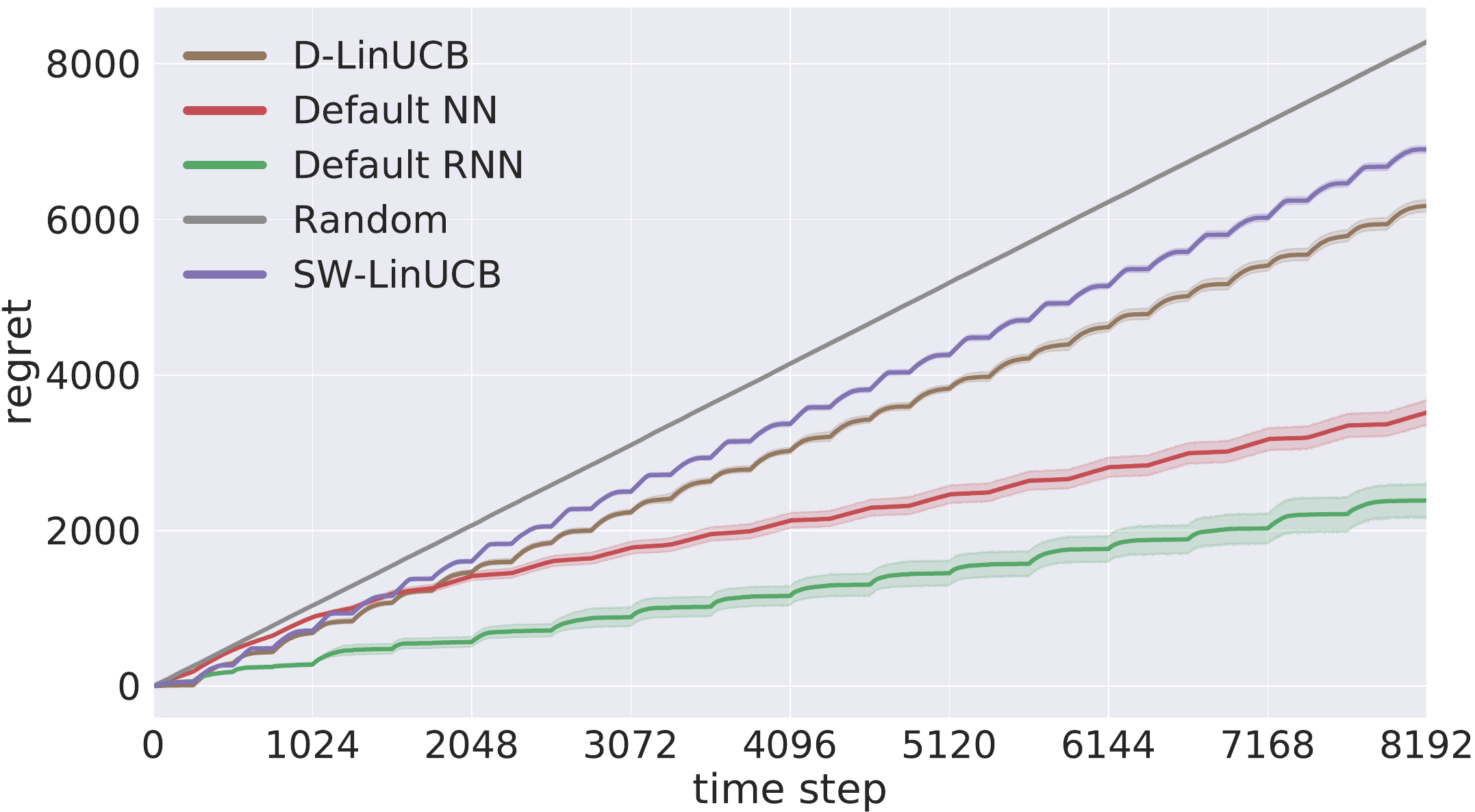}
  \caption{Flipping vector ($h=256$).}
  \label{fig:criteo_exps:periodic_50}
\end{subfigure}
\caption{Regret curves for non-stationary bandit problems with the Criteo dataset.}
\label{fig:criteo_exps}
\end{figure}

In the setting with the single abrupt partial flip, the conventional non-stationary bandit algorithms (SW/D-LinUCB) perform the best (Fig. \ref{fig:criteo_exps:abrupt_50}). The default recurrent policy performs admirably well considering that the conventional approaches are highly suited to this problem. Unlike the RNN-based approach, SW-LinUCB and D-LinUCB are equipped with the inductive bias of a linear model that matches the problem. Furthermore, they benefit from the additional knowledge of the variation budget, which is used to set their parameters (window size and discount) according to what minimizes their theoretical regret. Note that the default hyperparameters of the neural policies are not tuned for this task.

When we increase the variation budget by constructing an experiment with periodic abrupt flips every $h$ steps (Fig. \ref{fig:criteo_exps:periodic_50}), we observe that the recurrent policy is able to exploit the periodicity to outperform the other approaches. This effect is similar to the one seen in the faster rotating vector problem.

\textbf{Stationary bandit problems.} In these experiments, we present regret curves for the default policies in stationary variants of the flipping Bernoulli and flipping vector problems ($h \to \infty$, see Section \ref{sec:experiments:bandit_problems:stationary}).

\begin{figure}[ht]
    \centering
    \begin{floatrow}
      \ffigbox{\caption{Stationary Bernoulli.}\label{fig:stationary_bernoulli_main}}{%
        \includegraphics[width=1.0\linewidth]{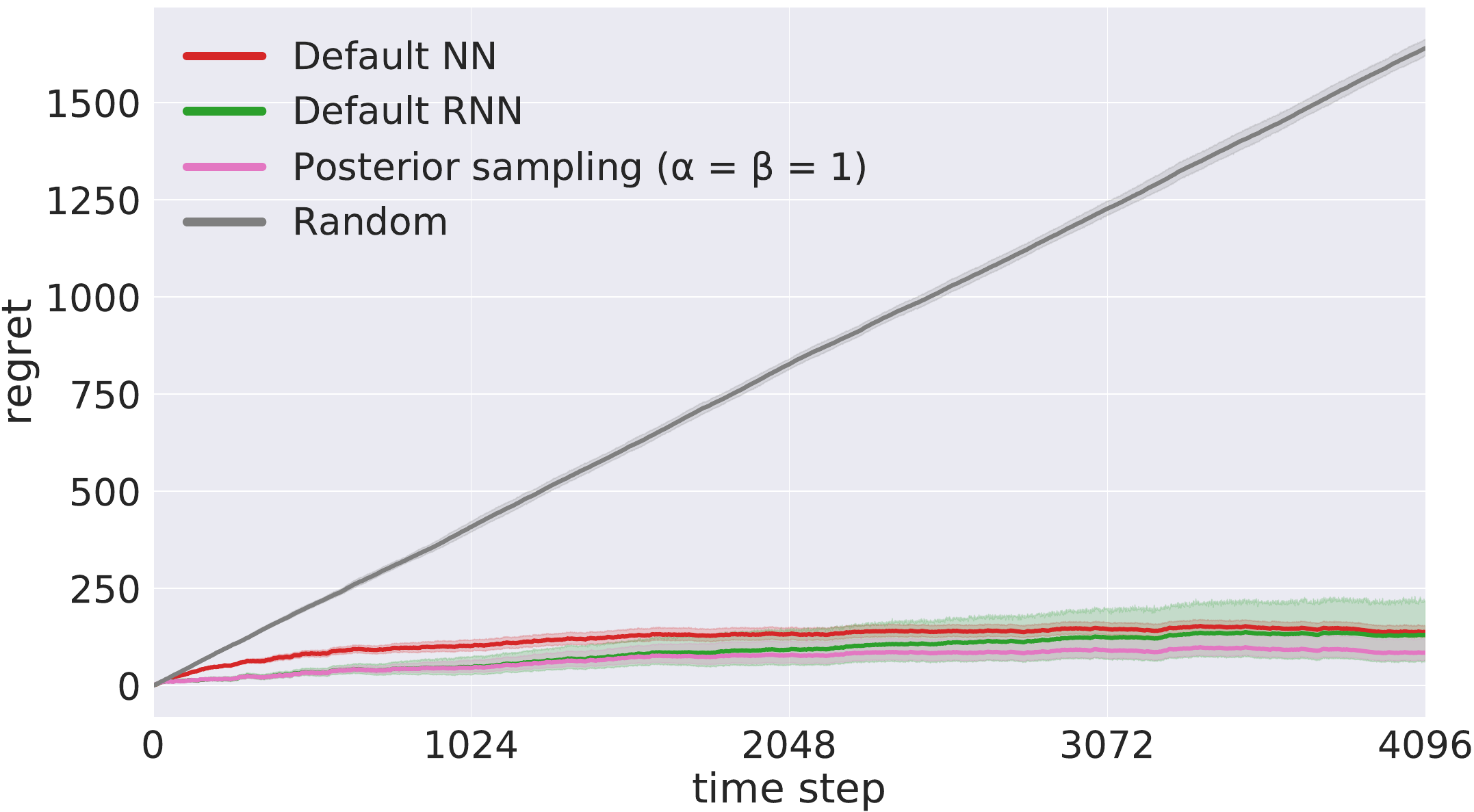}
      }
    \ffigbox{\caption{Stationary vector.}\label{fig:stationary_vector_main}}{%
        \includegraphics[width=1.0\linewidth]{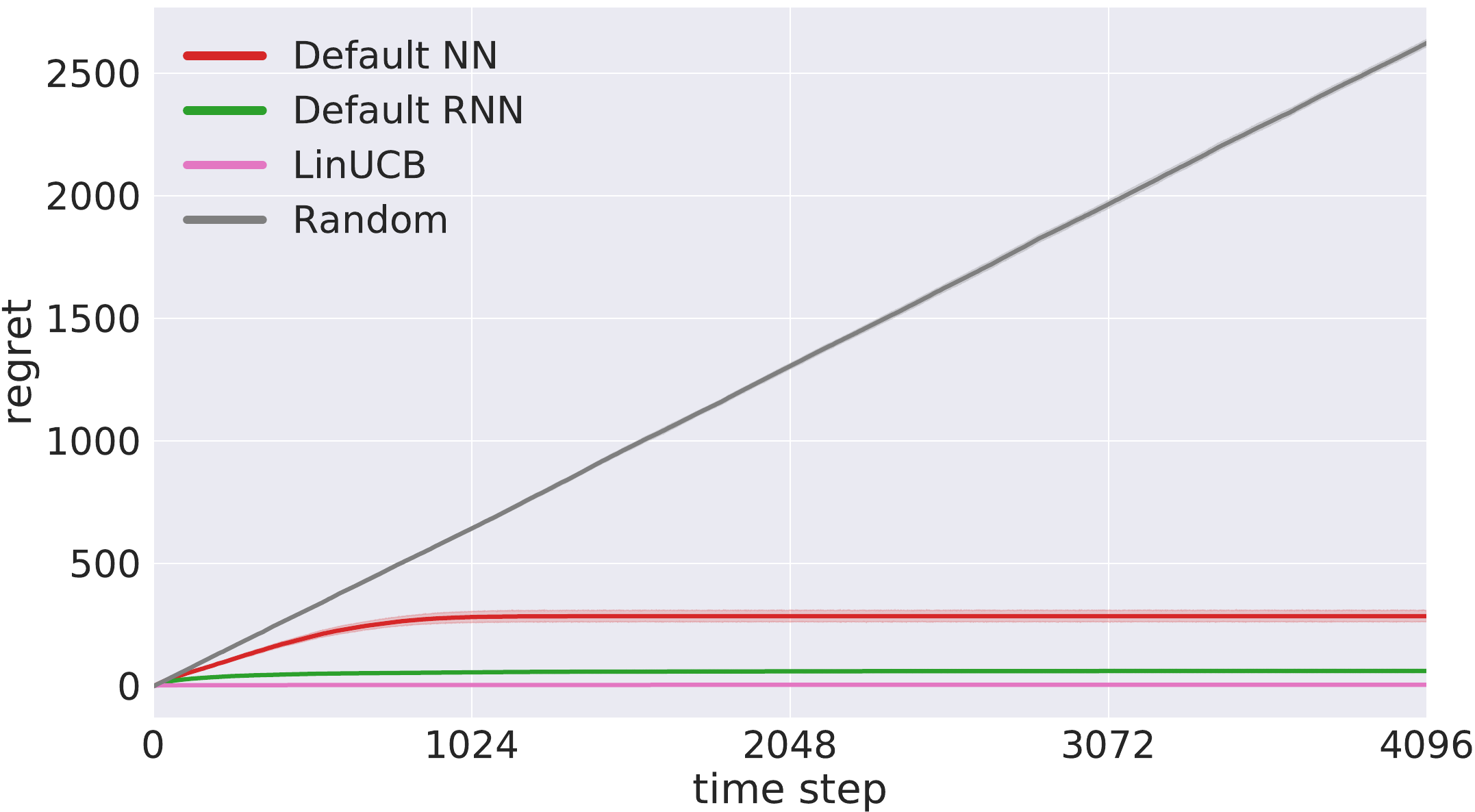}
      }
    \end{floatrow}
\end{figure}

In the stationary Bernoulli problem (Fig. \ref{fig:stationary_bernoulli_main}), the regret curves of the default (R)NN policies are comparable to the curve for the conventional posterior sampling policy for Bernoulli bandits \citep{thompson1933likelihood}. In the stationary vector problem (Fig. \ref{fig:stationary_vector_main}), the default NN is outperformed by the default RNN, which achieves a performance that is comparable to LinUCB.

These experiments indicate that the recurrent neural-linear approach is also capable of performing well in stationary problems.

\textbf{Alternatives to posterior sampling.} %
Finally, we present the results of an investigation into whether posterior sampling excels in comparison with simpler exploration strategies such as $\epsilon$-greedy exploration and Boltzmann exploration, which can also be combined with recurrent neural-linear features.

We compare posterior sampling with these two alternative strategies in two non-stationary bandit problems (flipping Gaussian and sinusoidal Bernoulli). For each problem, we use the same experimental protocol as before to find the best hyperparameters for each strategy (see Section \ref{sec:evaluation}). The hyperparameter grid detailed in Table \ref{tab:hgrid_defaults} is appropriately adapted to consider the $\epsilon$-greedy exploration factors $\epsilon \in \{ 0.01, 0.03, 0.05, 0.1 \}$ and Boltzmann exploration temperatures $\tau_k \in \{ 0.001, 0.005, 0.01, 0.05 \}$.

\begin{figure}[ht]
\begin{subfigure}{.49\linewidth}
  \centering
  \includegraphics[width=1.0\linewidth]{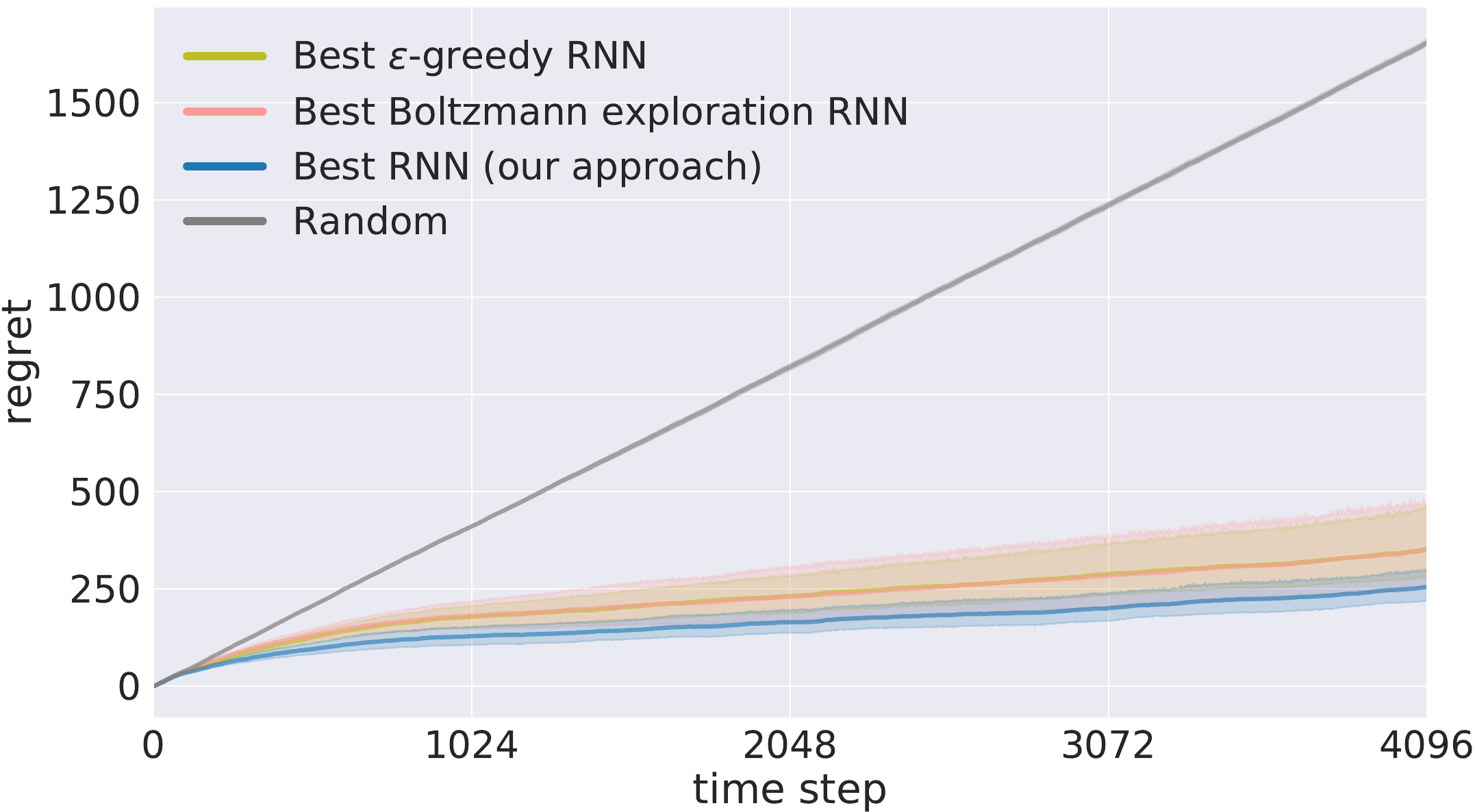} 
  \caption{Flipping Gaussian.}
  \label{fig:exploration_comparison:flip_gaussian}
\end{subfigure}
\begin{subfigure}{.49\linewidth}
  \centering
  \includegraphics[width=1.0\linewidth]{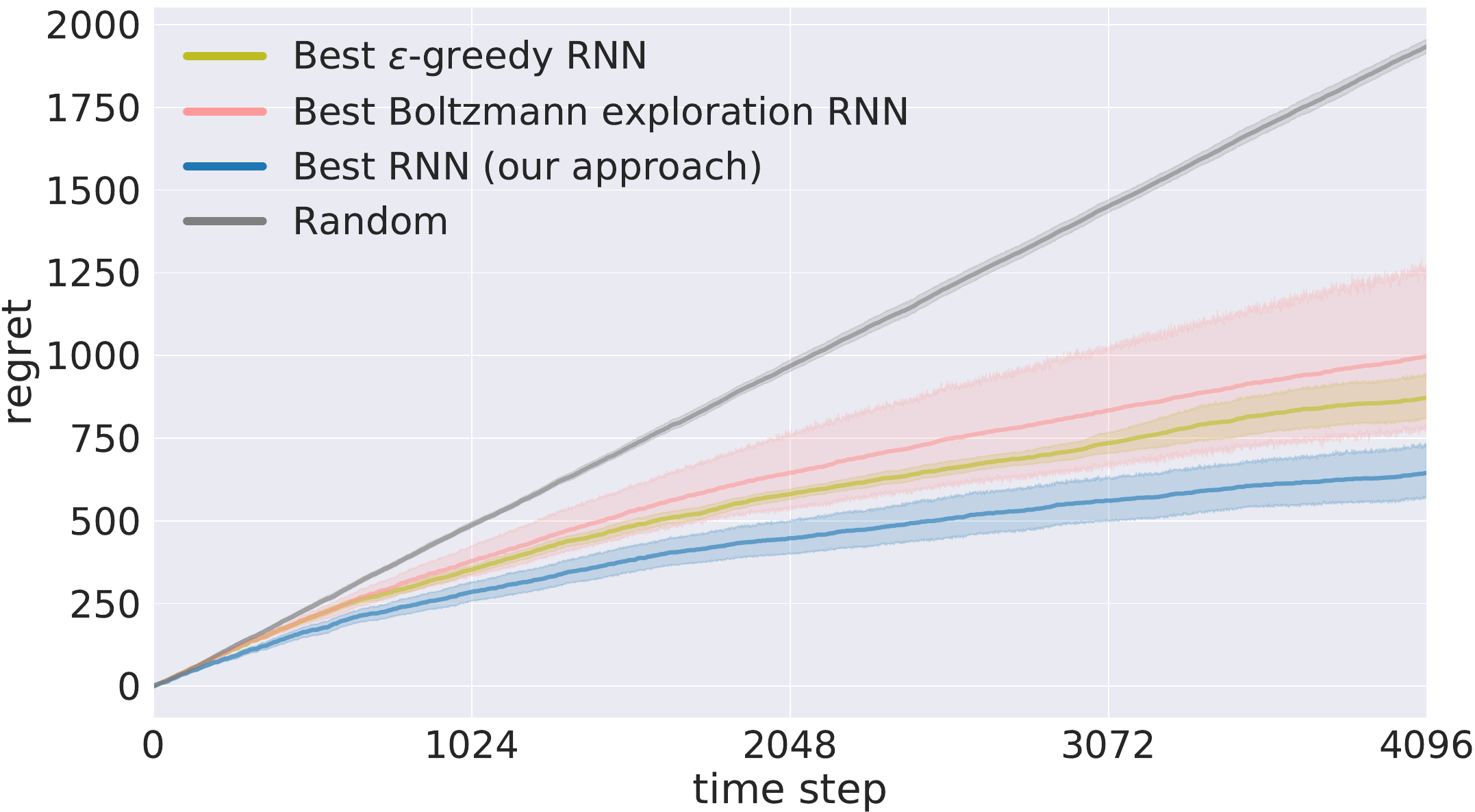} 
  \caption{Sinusoidal Bernoulli.}
  \label{fig:exploration_comparison:sin_bernoulli}
\end{subfigure}
\caption{Comparison between posterior sampling (our approach), $\epsilon$-greedy exploration, and Boltzmann exploration.}
\label{fig:exploration_comparison}
\end{figure}

The corresponding regret curves are presented in Fig. \ref{fig:exploration_comparison}. Posterior sampling outperforms the alternative approaches in both problems, which agrees with the results obtained by \citet{riquelme2018deep} for stationary problems.

\section{Conclusion}
\label{sec:conclusion}

We introduced an approach to non-stationary contextual bandit problems that learns to represent the relevant context for a decision based solely on the raw history of interactions between the agent and the environment. Prior to our work, solving such problems required carefully handcrafting a historical context, which could introduce spurious relationships or omit a convenient representation of crucial information; or employing conventional non-stationary bandit algorithms, whose assumptions had to coincide with the (typically unknown) underlying changes to the environment. Notably, our approach is also radically different from previous approaches that are able to exploit periodic or structured patterns in non-contextual non-stationary bandit problems.

The success of our approach relies on the strong assumption that the expected reward for any action at a given time step can be predicted as a (fixed but unknown) linear function of features extracted by a recurrent neural network that was trained to predict previous rewards.  Consequently, it is difficult to provide theoretical %
guarantees comparable to those provided by conventional bandit algorithms, which is the most significant drawback of our approach. 
Nevertheless, our novel analysis for linear posterior sampling with measurement error may serve as a foundation for future theoretical work.

Our experiments on a diverse selection of %
non-stationary bandit problems show that our approach achieves satisfactory performance on problems that can be solved by conventional non-stationary bandit algorithms, while also being applicable when such algorithms fail completely. Our approach also consistently outperforms its feedforward counterpart, which requires handcrafting a historical context, even when its hyperparameters are fixed across very dissimilar environments. These findings make our approach particularly appealing when there is limited knowledge about a problem. 

Another potential weakness of our approach is the computational cost of backpropagation through time, which is required to train the recurrent neural network. Fortunately, this issue may be mitigated by reducing the frequency of network training steps (as we have done), or by employing truncated backpropagation through time. These alternatives may compromise the quality of the learned contexts.

Because there are no standard benchmarks for non-stationary contextual bandit algorithms, we employed our own selection of problems, some of which were borrowed from previous work. Future work could focus on finding, creating, and adapting problems to further evaluate our approach. We are particularly interested in realistic applications and adversarial (adaptive) environments.

There are many possibilities for future work besides integrating our approach into real applications: combining alternative Bayesian recurrent neural network approaches with posterior sampling; combining recurrent neural-linear features with other contextual linear bandit algorithms; designing specialized recurrent architectures; improving recurrent neural network parameter initialization; inferring the variance of the reward distribution; providing theoretical guarantees for restricted classes of problems; and comparing our approach with additional non-stationary contextual bandit algorithms.

\subsection*{Acknowledgments}
We would like to thank Sjoerd van Steenkiste, Claire Vernade, Anand Gopalakrishnan, Francesco Faccio, Roshan Shariff, Dylan Ashley, Raoul Malm, and Imanol Schlag for their valuable feedback. This research was supported by Swiss Natural Science Foundation grants (200021\_165675/1 and 200021\_192356) and an ERC Advanced Grant (742870).

\bibliographystyle{apalike}
\bibliography{bib}

\clearpage
\section*{Appendix}
\appendix

\section{Linear posterior sampling with measurement error}
\label{app:theoretical_analysis}

This appendix presents the details of the proof of Theorem \ref{th:regret}, which introduces a new cumulative regret bound for linear posterior sampling with measurement error.

Section \ref{sec:theoretical_analysis:preliminaries} introduced the linear contextual bandit with measurement error setting. Appendix \ref{app:theoretical_analysis:regret_theorem} reproduces the proof of Theorem \ref{th:regret} which was provided in Section \ref{sec:theoretical_analysis:theorem} with additional details on the asymptotic analysis. Appendices  \ref{app:theoretical_analysis:lemma1}, \ref{app:theoretical_analysis:lemma2}, \ref{app:theoretical_analysis:lemma3}, and
\ref{app:theoretical_analysis:lemma4}
contain the key Lemmas required to prove Theorem \ref{th:regret}.

Our proof also relies on the following definitions and lemmas which are reported for completeness.

\smallskip
\begin{definition}[Martingale process] \label{definition:martingale}
A stochastic process $\left (Y_t; t\geq 0 \right )$ corresponding to filtration $\fil_t$ with $Y_t$ $\fil_t$-measurable, is called 
martingale w.r.t. $\fil_t$ if
$\,
\eb(Y_t - Y_{t-1} \mid \fil_{t-1}) = 0,
$
super-martingale w.r.t. $\fil_t$ if
$\,
\eb(Y_t - Y_{t-1} \mid \fil_{t-1}) \leq 0
$, and
sub-martingale w.r.t. $\fil_t$ if
$\,
\eb(Y_t - Y_{t-1} \mid \fil_{t-1}) \leq 0.
$
\end{definition}
\smallskip
\begin{definition}[Sub-Gaussianity] \label{definition:subG}
A random variable X is $\sigma$-sub-Gaussian if, for all $\lambda \in \rb$,
$$ \mathbb{E} \left ( \exp{ \left( \lambda X \right)} \right ) \leq \exp{\left( \frac{-\lambda^2 \sigma^2}{2} \right)}.$$
\end{definition}
\smallskip
\begin{lemma}[Azuma-Hoeffding inequality] \label{lemma:azuma_hoeffding}
If a super-martingale ($Y_t; t\geq0$) corresponding to filtration $\fil_t$ satisfies $|Y_t - Y_{t-1}| \leq c_t$ for some constant $c_t$, for all $t \in [T]$, then, for any $a \in \rb^+$,
\begin{equation}
    P \left (Y_T - Y_0 \geq a \right ) \leq \exp\left(-\frac{a^2}{2\sum_{t=1}^Tc_t^2}\right).
\end{equation}
\end{lemma}
\smallskip
\begin{lemma}[Lemma 8 by \citet{agrawal2012thompson}] \label{lemma:abba}
Let $(\fil'_t; t\geq0)$ be a filtration, $(\mathbf{m}_t; t\geq1)$ be a $\rb^d$-valued stochastic process such that $\mathbf{m}_t$ is $\fil'_{t-1}$-measurable, $(\eta_t; t\geq1)$ be a $\rb$-valued martingale process (see Definition~\ref{definition:martingale}) such that $\eta_t$ is $\fil_t$-measurable.
For $t\geq 0$, let
\begin{equation}
     \mathbf{\xi}_t = \sum_{i=1}^t \eta_i \mathbf{m}_i
     \quad \text{and} \quad
     \mathbf{M}_t = I_d + \sum_{i=1}^t \mathbf{m}_i \otimes \mathbf{m}_i.
\end{equation}
Assuming that $\eta_t$ is conditionally $\sigma$-sub-Gaussian,
\begin{equation}
\|\mathbf{\xi}_t\|_{\mathbf{M}^{-1}_t} \leq \sigma \sqrt{d\log\left(\frac{t+1}{\delta'}\right)}
\end{equation}
with probability $1-\delta'$.
\end{lemma}
\smallskip
\begin{lemma}[Elliptical potential lemma \citep{carpentier2020elliptical, lattimore19bandit, cesa2006prediction}] \label{lemma:epl}
Let $\mathbf{u}_1, \mathbf{u}_2, \dots, \mathbf{u}_T$ be any sequence of vectors in $\rb^d$ such that $\|\mathbf{u}_i\| \leq 1$ for all $i \in [T]$.
For any $t \in [T]$, let
\begin{equation}
     \mathbf{V}_t = \lambda I_d + \sum_{i=1}^{t-1} \mathbf{u}_i \otimes \mathbf{u}_i.
\end{equation}
Then, we have
\begin{equation}
\sum_{t=1}^{T}\|\mathbf{u}_t\|_{\mathbf{V}^{-1}_{t+1}} \leq \sqrt{T d\log\left(\frac{T + d \lambda}{d \lambda}\right)}.
\end{equation}
\end{lemma}

\subsection{Regret of linear posterior sampling with measurement error} \label{app:theoretical_analysis:regret_theorem}
\theoremregret*

\textit{Proof outline.}\hspace{1mm}
Under the assumption that $\ewt$ holds,
we construct a stochastic process $\big(Y_t; t\geq0 \big)$ related to the cumulative regret.
By Lemma~\ref{th:lemma4}, $Y_t$ is a super-martingale, and so we can use the Azuma-Hoeffding inequality (see Def. \ref{lemma:azuma_hoeffding}) to bound it with high probability.
Finally, we eliminate the dependency on $\ewt$ by taking a union bound over the high probability bound from the Azuma-Hoeffding inequality and Lemma~\ref{th:lemma1}, which bounds the probability of the event $\ewt$.

\begin{proof}
Let $\big(Y_t = \sum_{i=1}^t Z_i; t\geq0 \big)$ be a stochastic process corresponding to $\fil_t$ where
\begin{equation}
    \texttt{regret}'_{t,A_t} = \texttt{regret}_{t,A_t} \mathcal{I} \left \{\ewt \right \}
    \hspace{.2cm}
    \text{and}
    \hspace{.2cm}
    Z_t = \texttt{regret}'_{t,A_t} - 4\epsilon - 44 e \sqrt{\pi}g_t s_{t,A_t} - \frac{2}{t^2}.
\end{equation}
By Lemma~\ref{th:lemma4}, $Y_t$ forms a super-martingale process as
\begin{equation*}
   \eb \left( Y_t - Y_{t-1} \right) = \eb \left(Z_t\right) \leq 4 + 4\epsilon + 44e\sqrt{\pi}g_T = K_T.
\end{equation*}
Using the Azuma-Hoeffding inequality for super-martingale processes,
\begin{equation}\label{eq:t2_app}
\frac{Y_T - Y_0}{K_T} \leq \sqrt{2T\log \left ( \frac{2}{\delta} \right )}  
\end{equation}
with $1-\frac{\delta}{2}$ probability.

Taking the union bound over Lemma~\ref{th:lemma1} and \eq{t2_app}, we can bound the regret as
\begin{align} \label{eq:regret_final_app}
\sum_{t=1}^T \texttt{regret}_{t,A_t} \leq & 4\epsilon T + 44 e \sqrt{\pi}g_T 5\sqrt{dT\log T} + \frac{\pi^2}{3} \notag \\ 
& + \left (4 + 4\epsilon + 44e\sqrt{\pi}g_T \right )\sqrt{2T\log \left ( \frac{2}{\delta} \right )}
\end{align}
 with $1-\delta$ probability.
 
We now proceed to the asymptotic analysis of the regret bound in \eq{regret_final_app}, which concludes the proof.
Note that the second term of \eq{regret_final_app} dominates over the other terms.

As $g_T$ contains $\min \{\sqrt{d\log(T)},\sqrt{\log(TK)}\}$ in its definition, we analyze each of the two cases separately.
Intuitively, the first case happens when the number of actions $K$ is significantly larger than the dimensionality ($\sqrt{d\log(T)} < \sqrt{\log(TK)}$).
This result would also hold for $K=\infty$.
The second happens when the dimensionality dominates over the number of actions ($\sqrt{\log(TK)} < \sqrt{d\log(T)}$).

\paragraph{Case: $\sqrt{d\log(T)} < \sqrt{\log(TK)}$.}
The regret is in $\widetilde{O} \left( d^{\frac{3}{2}} \sqrt{T}  + \epsilon d^{\frac{3}{2}} T \right)$.
Let $L = 44 \cdot 5 \cdot e \sqrt{\pi}$ and $\,\widehat{R} = L g_T \sqrt{dT\log T}$. We have
\begin{align} \label{eq:case1.11}
g_T =& 2\sqrt{d\log(T)} \cdot \left ( 3 \sigma \sqrt{d\log\left(\frac{T}{\delta}\right)} + \epsilon \sqrt{T d \log \left ( \frac{T+d}{d} \right ) }\right ) \notag \\
& + \sigma \sqrt{d\log\left(\frac{T^3}{\delta}\right)}+ 1 + \epsilon \sqrt{T d \log \left ( \frac{T+d}{d} \right ) }.
\end{align}
The second term in \eq{case1.11} is asymptotically smaller, so we focus on the first term. Ignoring constant factors, we have
\begin{equation}
\widehat{R} \approx d \log(T) \sqrt{T} \cdot \left ( \sqrt{d \log \left( \frac{T}{\delta} \right) } + \epsilon \sqrt{T d \log \left ( \frac{T+d}{d} \right ) } \right),
\end{equation}
which gives us that the regret is in $\widetilde{O} \left( d^{\frac{3}{2}} \sqrt{T}  + \epsilon d^{\frac{3}{2}} T \right)$, where $\widetilde{O}$ ignores polylog factors.

\paragraph{Case: $\sqrt{\log(TK)} < \sqrt{d\log(T)}$.}
The regret is in $\widetilde{O} \left( d\sqrt{T \log(K)} + \epsilon \sqrt{d \log(K)} T^{\frac{3}{2}} \right)$.
In this case, we have
\begin{align} \label{eq:case1.1}
g_T =& 2\sqrt{\log(TK)} \cdot \left ( 3 \sigma \sqrt{d\log\left(\frac{T}{\delta}\right)} + \epsilon \sqrt{T d \log \left ( \frac{T+d}{d} \right ) } \right ) \notag \\  
& + \sigma \sqrt{d\log\left(\frac{T^3}{\delta}\right)} + 1 + \epsilon \sqrt{T d \log \left ( \frac{T+d}{d} \right ) }.
\intertext{For large $T$ and $K$, this yields}
g_T \leq & 2\sqrt{\log(T) \log(K)} \cdot \left ( 3 \sigma \sqrt{d\log\left(\frac{T}{\delta}\right)} + \epsilon \sqrt{T d \log \left ( \frac{T+d}{d} \right ) } \right ) \notag \\
& + \sigma \sqrt{d\log\left(\frac{T^3}{\delta}\right)}+ 1 + \epsilon \sqrt{T d \log \left ( \frac{T+d}{d} \right ) }.
\end{align}
Considering only the dominant term, which is the first, we have
\begin{equation}
 \widehat{R} \approx \log(T) \sqrt{dT \log(K)} \cdot \left ( \sqrt{d \log \left( \frac{T}{\delta} \right) } + \epsilon \sqrt{T d \log \left ( \frac{T+d}{d} \right ) } \right),
\end{equation}
which gives us the regret in $\widetilde{O} \left( d\sqrt{T \log(K)} + \epsilon d\sqrt{\log(K)} T \right)$.

By combining the two cases, the resulting regret bound is in
\begin{equation}
\widetilde{O} \left(
d^{\frac{3}{2}} \sqrt{T} + \epsilon d^{\frac{3}{2}} T
\right)
\quad
\text{or}
\quad
\widetilde{O}
\left(
d\sqrt{T \log(K)} + \epsilon d \sqrt{\log(K)} T
\right),
\end{equation}
whichever is smaller, with probability $1-\delta$.
\end{proof}

\subsection{Lemma \ref{th:lemma1}:} \label{app:theoretical_analysis:lemma1}

\begin{lemma}[Bounding the probability of events $\ewt$ and $\ett$] \label{th:lemma1}
For all $t \geq 0$, $0 < \delta < 1$, 
\begin{equation}
    P(\ewt) \geq 1  - \frac{\delta}{t^2}
    \hspace{0.4cm}
    \text{and}
    \hspace{0.4cm}
    P(E^{\Theta}_t\mid \fil_{t-1}) \geq 1 - \frac{1}{t^2}.
\end{equation}
\end{lemma}

\begin{proof}

For the proof of $\ett$, we refer to the proof of Lemma~1 by \citet{agrawal2012thompson} since, by construction, the event $\ett$ is a concentration around the mean event for Gaussian random variables.

For $\ewt$, instead, the proof is substantially different due to the presence of errors-in-variables.

We use Lemma \ref{lemma:abba}, setting $\mathbf{m}_t = \pst$ and $\eta_t = r_{t,A_t} - \pht\cdot\mathbf{w}$ which results in $\mathbf{M}_t = \mathbf{B}_{t+1}$ and ${\mathbf{\xi}_t = \sum_{i=1}^t (r_{i,a_i} - \phs\cdot\mathbf{w}) \pss}$.
By definition of the filtration, $\mathbf{m}_t$ is $\fil_{t-1}$-measurable and $\eta_t$ is $\fil_t$-measurable.
Since $\eta_t$ is also assumed to be conditionally $\sigma$-sub-Gaussian,
$(\eta_t, t\geq0)$ forms a martingale process.

In the following, we prove an upper bound for ${|\psk \cdot \whct - \phk \cdot \mathbf{w}|}$ in terms of $\mathbf{\xi}_t$ that allows us to use Lemma \ref{lemma:abba}.\\
Define the matrix forms of previous contexts as $\mathbf X_t =
\begin{bmatrix}
\mathbf{x}_{1,a_1}^T\\
\vdots\\
\mathbf{x}_{t-1,a_{t-1}}^T
\end{bmatrix}
$
and $\mathbf X^*_t =
\begin{bmatrix}
\mathbf{x^*}_{1,a_1}^T\\
\vdots\\
\mathbf{x^*}_{t-1,a_{t-1}}^T
\end{bmatrix}
$.
\begin{align}
    \intertext{Writing $\mathbf{\xi}_{t-1}$ in matrix form, $\mathbf X_t^T \mathbf r_t - \mathbf X_t^T  \mathbf X_t^* \mathbf{w}$, we obtain through algebraic manipulations that}
    &\whct - \mathbf{w} = \mathbf{B}_t^{-1}\left (\mathbf{\xi}_{t-1} - \left (\mathbf{B}_t - \mathbf X_t^T  \mathbf X_t^* \right) \mathbf{w} \right). \label{eq:p1.1}
\end{align}

We now have all the elements to start the proof.
\begin{align}
\intertext{By a sequence of applications of the \cs inequality,}
    |\psk \cdot \whct - \phk \cdot \mathbf{w}| \leq\,&
    |\psk \cdot (\whct - \mathbf{w})| + \|\psk - \phk\|\|\mathbf{w}\|.
\intertext{
Since $\|\mathbf{w}\|\leq1$ and $\|\psk - \phk\| \leq \epsilon$,
}
|\psk \cdot \whct - \phk \cdot \mathbf{w}|\leq\,&|\psk \cdot (\whct - \mathbf{w})| + \epsilon.
\intertext{Combining \eq{p1.1} with the \cs inequality and the definition of $s_{t,k}$,}
|\psk \cdot \whct - \phk \cdot \mathbf{w}|\leq\,& s_{t,k} \, \|\mathbf{\xi}_{t-1} - (\mathbf{B}_t - \mathbf X_t^T  \mathbf X_t^*) \mathbf{w}\|_{\mathbf{B}_t^{-1}} + \epsilon.  \label{eq:l1.t}
\end{align}

Note that by the \cs inequality and Lemma~\ref{lemma:abba},
\begin{align}
    \|\mathbf{\xi}_{t-1} - \left (\mathbf{B}_t - \mathbf X_t^T  \mathbf X_t^* \right) \mathbf{w}\|_{\mathbf{B}_t^{-1}}\leq\sigma \sqrt{d\log \left (\frac{t}{\delta'} \right)} + \|\left (\mathbf{B}_t - \mathbf X_t^T  \mathbf X_t^* \right) \mathbf{w}\|_{\mathbf{B}_t^{-1}} \label{eq:l1.s}
\end{align}
with probability $1-\delta'$.

It only remains to bound $\|(\mathbf{B}_t - \mathbf X_t^T  \mathbf X_t^*) \mathbf{w}\|_{\mathbf{B}_t^{-1}}$.
\begin{align}
\intertext{Using the definition of $\mathbf{B}_t$ and the \cs inequality,}
    \|(\mathbf{B}_t - \mathbf X_t^T  \mathbf X_t^*) \mathbf{w}\|_{\mathbf{B}_t^{-1}} \leq\,&
    \|\mathbf{w}\|_{\mathbf{B}_t^{-1}} + \|\mathbf X_t^T(\mathbf X_t - \mathbf X_t^*)\mathbf{w}\|_{\mathbf{B}_t^{-1}}.
\intertext{Using $\|\mathbf{w}\|\leq1$ and converting $\mathbf X_t^T  \mathbf X_t^*$ to summation form,}
   \|(\mathbf{B}_t - \mathbf X_t^T  \mathbf X_t^*) \mathbf{w}\|_{\mathbf{B}_t^{-1}} \leq\,& 1 + \|\sum_{i=1}^{t-1}\pss \otimes (\pss - \phs) \cdot \mathbf{w}\|_{\mathbf{B}_t^{-1}}.
\intertext{By applying the \cs inequality on the sum, and using the bounds on the measurement error and $\mathbf{w}$,}
   \|(\mathbf{B}_t - \mathbf X_t^T  \mathbf X_t^*) \mathbf{w}\|_{\mathbf{B}_t^{-1}} \leq\,& 1 + \epsilon \sum_{i=1}^{t-1}\| \pss\|_{\mathbf{B}_t^{-1}} \label{eq:l1.f}.
\end{align}
We can observe that the sequence of inverse covariance matrices $\mathbf{B}_1^{-1}, \mathbf{B}_2^{-1} \dots \mathbf{B}_t^{-1}$ is ``decreasing'' by construction. %
\begin{equation}
    1 + \epsilon \sum_{i=1}^{t-1}\| \pss\|_{\mathbf{B}_t^{-1}} \leq 1 + \epsilon \sum_{i=1}^{t-1}\| \pss\|_{\mathbf{B}_i^{-1}}
\end{equation}
Further applying the elliptical potential lemma (see Lemma \ref{lemma:epl}), we get 
\begin{equation}
    \|(\mathbf{B}_t - \mathbf X_t^T  \mathbf X_t^*) \mathbf{w}\|_{\mathbf{B}_t^{-1}} \leq 1 + \epsilon \sum_{i=1}^{t-1}\| \pss\|_{\mathbf{B}_i^{-1}} \leq 1 + \epsilon \sqrt{t d \log \left ( \frac{t+d}{d} \right ) }
\end{equation}

Combining Eqs. (\ref{eq:l1.t}), (\ref{eq:l1.s}), and (\ref{eq:l1.f}) and setting $\delta'=\frac{\delta}{t^2}$, we obtain that
\begin{equation}
|\psk \cdot \whct - \phk \cdot \mathbf{w}| 
\leq  
s_{t,k} \left (\sigma \sqrt{d\log\left(\frac{t^3}{\delta} \right) } + 1 + \epsilon \sqrt{t d \log \left ( \frac{t+d}{d} \right ) }\right)  + \epsilon
\end{equation}
happens with probability $1-\frac{\delta}{t^2}$ and conclude the proof for $\ewt$.
\end{proof}

\subsection{Lemma \ref{th:lemma2}:} \label{app:theoretical_analysis:lemma2}

\begin{lemma}[Lower bound for the probability that, for the optimal arm, the predicted mean exceeds the true mean] \label{th:lemma2}
For any filtration $\fil_{t-1}$ such that $\ewt$ is true,
\begin{equation}
    P \left (\Theta_{t,a_t^*} + \epsilon \geq \phtOpt \cdot \mathbf{w} \mid \fil_{t-1}, \ewt \right ) \geq (4e\sqrt{\pi})^{-1}. 
    \label{eq:lemma2}
\end{equation}
\end{lemma}

\begin{proof}
Recall that, for all $k \in [K]$, $\Theta_{t,k} = \wtct \cdot \psk \sim \N{\whct \cdot \psk}{\nu_t^2s_{t,k}^2}$.
\small
\begin{align}
\intertext
{
Subtracting the mean and dividing by the standard deviation of $\Theta_{t,a^*}$ in \eq{lemma2},
}
P\left (\Theta_{t,a_t^*} + \epsilon \geq \phtOpt \cdot \mathbf{w} \mid \fil_{t-1}, \ewt \right)
= & 
P \left (\frac{\Theta_{t,a_t^*} - \pstOpt \cdot \whct + \epsilon}{\nu_t s_{t,a_t^*}}
    \geq \frac{\phtOpt \cdot \mathbf{w} - \pstOpt \cdot \whct}{\nu_t s_{t,a_t^*}}
    \mid \fil_{t-1} \right).
\intertext
{
Applying the absolute value,
}
P \left (\Theta_{t,a_t^*} + \epsilon \geq \phtOpt \cdot \mathbf{w} \mid \fil_{t-1}, \ewt \right )
\geq&
P \left (\frac{\Theta_{t,a_t^*} - \pstOpt \cdot \whct}{\nu_t s_{t,a_t^*}} 
    \geq \frac{|\phtOpt \cdot \mathbf{w} - \pstOpt \cdot \whct| - \epsilon}{\nu_t s_{t,a_t^*}}
    \mid \fil_{t-1} \right).
\intertext
{
Finally, noting that $l_t s_{t, k} \leq \nu_t s_{t, k}$ and thus $\frac{|\phtOpt \cdot \mathbf{w} - \pstOpt \cdot \whct| - \epsilon}{\nu_t s_{t,a_t^*}}\leq 1$,
}
P \left (\Theta_{t,a_t^*} + \epsilon \geq \phtOpt \cdot \mathbf{w} \mid \fil_{t-1}, \ewt \right )
\geq&
P \left (Z \geq 1 \mid \fil_{t-1}, \ewt \right ),
\end{align}
\normalsize
where $Z \sim \mathcal{N}(0, 1)$.

The probability of $Z\geq1$ is lower bounded by $(4e\sqrt{\pi})^{-1}$ using an anti-concentration inequality for normally distributed random variables.
\end{proof}

\subsection{Lemma \ref{th:lemma3}:} \label{app:theoretical_analysis:lemma3}

\begin{lemma}[Bound for the probability of playing saturated arms in terms of the probability of playing unsaturated arms] \label{th:lemma3}

For any filtration $\fil_{t-1}$ such that $\ewt$ is true,
\begin{equation}
    P \left (a_t \notin C_t| \fil_{t-1}, \ewt \right ) \geq (4e\sqrt{\pi})^{-1} - \frac{1}{t^2}.
\end{equation}
\end{lemma}

\begin{proof}
Let $\ac = \left \{\forall k \in [K],\,\, \psk \cdot \wtct < \Theta_{t,a_t^*} \mid \fil_{t-1}, \ewt \right \}$ and $\mathcal{B}= \left \{a_t \notin C_t| \fil_{t-1}, \ewt \right \}$.
Since that action at time $t$ is chosen as $a_t = \arg\max_{a_t} \pst \cdot \wtct$, $\ac$ implies $\mathcal{B}$, so a lower bound for the probability of $\ac$ is a lower bound for the probability of $\mathcal{B}$ as well.

In order to get a lower bound on the probability of $\ac$, we first upper bound the absolute difference between the true reward mean and the sampled reward mean from the posterior using the definitions of the events $\ewt$ and $\ett$.
\begin{align}
\intertext{Suppose that both $\ewt$ and $\ett$ hold.  By their definitions and the \cs inequality,}
    &|\phk \cdot \mathbf{w} - \psk \cdot \wtct| = |\phk \cdot \mathbf{w} - \psk \cdot \wtct \pm \psk \cdot \whct| \leq \epsilon + s_{t,k} g_t. \label{eq:l3.1}
\intertext{Hence, for all $k \in [K]$,}
      &\psk \cdot \wtct - \phk \cdot \mathbf{w} \leq |\psk \cdot \wtct - \phk \cdot \mathbf{w}|\leq\epsilon + s_{t,k} g_t. \label{eq:a3.5}
\intertext{In particular, for all $k \in C_t$,}
     &\psk \cdot \wtct - \phk \cdot \mathbf{w} \leq \phtOpt \cdot \mathbf{w} - \phk \cdot \mathbf{w} - \epsilon\label{eq:a3.4},
\end{align}
which implies that $\psk \cdot \wtct \leq \phtOpt \cdot \mathbf{w} - \epsilon$.

Let $\mathcal{C} = \left \{\phtOpt \cdot \mathbf{w} - \epsilon < \Theta_{t,a^*} \mid \fil_{t-1}, \ewt \right \}$.
If both $\mathcal{C}$ and $\ett$ hold, then $\ac$ holds as well, so $\ett \cap \mathcal{C} \subseteq \ac$. This can be used to derive a lower bound on the probability of $\ac$:
\begin{equation}
    \mathcal{C} = \mathcal{C} \cap \left (\ett \cup \bar \ett \right ) \subseteq \left ( \mathcal{C} \cap \ett \right ) \cup \bar \ett \subseteq \ac \cup \bar \ett
    \,\, \Rightarrow \,\, 
    P(\ac) \geq P(\mathcal{C}) - P(\bar \ett).
\end{equation}
\begin{align}
    \intertext{Finally, using Lemma~\ref{th:lemma1} for the probability of $\bar \ett$ and Lemma~\ref{th:lemma2} for the probability of $\mathcal{C}$,}
    &P(\mathcal{B}) \geq P(\ac) \geq P(\mathcal{C}) - P(\bar \ett) \geq (4e\sqrt{\pi})^{-1} - t^{-2},
\end{align}
which concludes the proof.
\end{proof}

\subsection{Lemma \ref{th:lemma4}:} \label{app:theoretical_analysis:lemma4}

\begin{lemma}[Bound for the expectation of the regret at each round conditioned on $\ewt$] \label{th:lemma4}

For any filtration $\fil_{t-1}$ such that $\ewt$ is true,
\begin{equation}
    \eb \left (\phtOpt \cdot \mathbf{w} - \pht\cdot \mathbf{w} | \fil_{t-1}, \ewt \right ) \leq 4\epsilon + 44 e \sqrt{\pi}g_t\eb \left (s_{t,a_t}\mid \fil_{t-1}, \ewt \right ) + \frac{2}{t^2}.
\end{equation}
\end{lemma}

\begin{proof}
Let $\bar a_t  = \arg\min_{k \notin C_t} s_{t,k}$ denote the unsaturated arm with smallest variance.
Note that $\bar a_t$ is fixed given a filtration $\fil_{t-1}$, since both $s_{t,k}$ and $C_t$ are fixed given $\fil_{t-1}$.

We first lower bound the expected value of the variance $s_{t,k}$ of a generic arm $k$ with the minimal variance across all the arms $s_{t,\bar a_t}$.
\begin{align}
    \intertext{By the tower property of expectation, the fact that $s_{t,k}$ is always positive, Lemma \ref{th:lemma3}, and the definition of $\bar a_t$,}
    \eb (s_{t,a_t}\mid\fil_{t-1}, \ewt) \geq&\, \eb (s_{t,a_t}\mid\fil_{t-1}, \ewt, a \notin C_t)P(a_t \notin C_t)\geq s_{t,\bar a_t}((4e\sqrt{\pi})^{-1} - \frac{1}{t^2}). \label{eq:a4.0}
\end{align}
We now bound the instantaneous regret, $\phtOpt \cdot \mathbf{w} - \pht\cdot \mathbf{w}$.
\smallskip
\begin{align}
    \intertext{Using the result previously obtained in \eq{l3.1} and by the definition of saturated arms,}
    \phtOpt \cdot \mathbf{w} - \pht \cdot \mathbf{w} \leq&
    2\epsilon + s_{t,\bar a_t} + \phtBar\cdot\mathbf{w} - \pht\cdot \mathbf{w} \pm 
    \pstBar \cdot \wtct \pm  \pst \cdot \wtct.\\
    \intertext{Assuming $\ett$ and using \eq{l3.1} in combination with the fact that $\pst\cdot\wtct \geq \pstBar\cdot\wtct$, which holds by the $\argmax$ action selection,}
    \phtOpt \cdot \mathbf{w} - \pht \cdot \mathbf{w} \leq& 4\epsilon + g_t(2s_{t,\bar a_t} + s_{t,a_t}). \label{eq:a4.1}
\end{align}
We can now combine Eqs. (\ref{eq:a4.0}) and (\ref{eq:a4.1}) to finalize the proof.
\smallskip
\small
\begin{align}
    \intertext{First, by \eq{a4.1},}
    \eb \left (\phtOpt \cdot \mathbf{w} - \pht\cdot \mathbf{w} \mid \fil_{t-1}, \ewt \right) & \leq
    4\epsilon + g_t (2 s_{t,\bar a_t} + \eb(s_{t,a_t}\mid \fil_{t-1}, \ewt)) \notag \\
    & + \eb(\phtOpt \cdot \mathbf{w} - \pht\cdot \mathbf{w} \mid \fil_{t-1}, \ewt, \bar\ett)P(\bar\ett).
\intertext{Using \eq{a4.0}, Lemma \ref{th:lemma1}, and the fact that $|\phtOpt \cdot \mathbf{w} - \pht\cdot \mathbf{w}|\leq 2$,}
    \eb(\phtOpt \cdot \mathbf{w} - \pht\cdot \mathbf{w} \mid \fil_{t-1}, \ewt)
    \leq&\,
    4\epsilon + g_t\left (\frac{2}{(4e\sqrt{\pi})^{-1}-t^2} + 1 \right )\eb(s_{t,a_t}\mid \fil_{t-1}, \ewt) \notag \\ &+\frac{2}{t^2}.
\intertext{Finally, since $\frac{t^2}{(4e\sqrt{\pi})^{-1}-t^2} \leq 20e\sqrt{\pi}$,}
    \eb \left (\phtOpt \cdot \mathbf{w} - \pht\cdot \mathbf{w} \mid \fil_{t-1}, \ewt \right )
    \leq&\,
    4\epsilon + 44e\sqrt{\pi}g_t\eb(s_{t,a_t}\mid \fil_{t-1}, \ewt) + \frac{2}{t^2},
\end{align}
\normalsize
which concludes the proof.
\end{proof}
\end{document}